\documentclass[nopreprintline,5p,times]{elsarticle}

\usepackage{amssymb}
\usepackage{amsmath}

\usepackage{booktabs}
\usepackage{color,soul}
\usepackage{caption}
\usepackage{subfigure}
\usepackage{siunitx}
\usepackage{soul}

\journal{Medical Image Analysis}

\begin{document}

\begin{frontmatter}



\title{FourierLoss: Shape-Aware Loss Function with Fourier Descriptors}

\author[label1,label2]{Mehmet Bahadir Erden}
\author[label1,label2]{Selahattin Cansiz}
\author[label1,label2]{Onur Caki}
\author[label4]{Haya Khattak}
\author[label5]{Durmus Etiz}
\author[label5]{Melek Cosar Yakar}
\author[label5]{Kerem Duruer}
\author[label5]{Berke Barut}
\author[label1,label2,label3]{Cigdem Gunduz-Demir\corref{cor1}}
\ead{cgunduz@ku.edu.tr}
\cortext[cor1]{Corresponding author.}

\affiliation[label1]{organization={Department of Computer Engineering, Koc University},
            postcode = {34450},
            city={Istanbul},
            country={Turkey}}

\affiliation[label2]{organization={KUIS AI Center, Koc University},
            postcode = {34450},
            city={Istanbul},
            country={Turkey}}

\affiliation[label3]{organization={School of Medicine, Koc University},
            postcode = {34010},
            city={Istanbul},
            country={Turkey}}
            
\affiliation[label4]{organization={Department of Computer Engineering, Bilkent University},
            postcode = {06800},
            city={Ankara},
            country={Turkey}}
            
\affiliation[label5]{organization={School of Medicine, Osmangazi University},
            postcode = {26040},
            city={Eskisehir},
            country={Turkey}}



\begin{abstract}
Encoder-decoder networks become a popular choice for various medical image segmentation tasks. When they are trained with a standard loss function, these networks are not explicitly enforced to preserve the shape integrity of an object in an image. However, this ability of the network is important to obtain more accurate results, especially when there is a low-contrast difference between the object and its surroundings. In response to this issue, this work introduces a new shape-aware loss function, which we name \textit{FourierLoss}. This loss function relies on quantifying the shape dissimilarity between the ground truth and the predicted segmentation maps through the Fourier descriptors calculated on their objects, and penalizing this dissimilarity in network training. Different than the previous studies, \textit{FourierLoss} offers an adaptive loss function with trainable hyperparameters that control the importance of the level of the shape details that the network is enforced to learn in the training process. This control is achieved by the proposed adaptive loss update mechanism, which end-to-end learns the hyperparameters simultaneously with the network weights by backpropagation. As a result of using this mechanism, the network can dynamically change its attention from learning the general outline of an object to learning the details of its contour points, or vice versa, in different training epochs. Working on 2879 computed tomography images of 93 subjects, our experiments revealed that the proposed adaptive shape-aware loss function led to statistically significantly better results for liver segmentation, compared to its counterparts.
\end{abstract}




\end{frontmatter}


\section{Introduction}

Organ segmentation in computed tomography (CT) scans is essential in the diagnosis of diseases and treatment planning. Applications such as the detection of anomalies and assessment of treatment responses require accurate segmentation of organs as a first and integral step. However, manual segmentation of an organ in numerous axial slices, containing the organ of interest in varying contrast and sizes, is time consuming and prone to observer variability. This problem becomes more challenging for soft tissues and organs due to the close reflectivity level of these tissues and organs and their surroundings. Therefore, many methods have been proposed for automatic segmentation of organs. Convolutional neural network-based architectures were widely used for organ segmentation in radiologic images. These networks consisted of symmetric encoder and decoder paths with long skip connections~\cite{li2018h, ronneberger2015u}. However, due to the design of their architectures, the predictions of such networks mainly depend on capturing local patterns and may fail to make effective use of the global context of a segmentation task. As a result, conventional networks may produce coarse and unsmooth segmentations.

One effective way of attending the global context of the segmentation task is to model the shape of the objects to be segmented and incorporate it into network training. This has been achieved either by defining a shape-preserving loss function or by defining shape-related auxiliary tasks in a multitask learning framework. Although the literature includes many studies that improved segmentations using these approaches, they may not fully exploit the shape information due to the use of coarse metrics/means to quantify the shape in the network design. For example, in~\cite{alarifShapeAwareDeepConvolutional2018}, the authors used the mean Euclidean distance between the boundaries of the ground truth and segmented regions as a penalty term in their loss function. However, averaging pixel-wise distances led to a loss in the representation of global shape information, and thus, provided a coarse measure of dissimilarity. In another study~\cite{chen2019learning}, the contour length and area of the objects were used. Similarly, these metrics were not distinct features for shape quantification since these features may be the same for objects of different shapes. It was also proposed to employ predefined filters or class-based probability maps as shape priors in network training~\cite{tofighi2019prior, zotti2018convolutional}. Since these priors represented the common characteristics of the training set, this use did not take account of specific shape-related details of individual objects.

Alternatively, distance transforms were used to generate maps that approximated the shape of the objects in the ground truths. Multitask networks were designed to learn the prediction of these maps along with the main task of segmentation~\cite{murugesan2019psi, wang2020deep}. Since these networks had a separate decoder path for each task, they resulted in a significant increase in the network parameters (weights) for the auxiliary task of predicting the distance transform map, which indeed could be inferred from the ground truth itself, and hence, from the segmentation task. In~\cite{karimi2019reducing}, the authors approximated the one-sided Hausdorff distance from a distance map of the ground truth objects and added it as a penalty term in the loss function. Nevertheless, the Hausdorff distance relies on measuring the maximum shape deviation from a ground truth object, and as a result, may not model fine details about the shape. Furthermore, all of these networks defined their shape preserving loss functions or auxiliary tasks at the beginning and the shape criteria enforced by these functions/tasks remained unchanged throughout training. As a result, the network cannot adaptively change the granularity of the shape it needs to learn. In other words, it cannot switch from learning the coarse shape details to the finer ones, or vice versa, during training. 

To address this gap, in this work, we propose a new shape-aware loss function that implicitly forces a network to learn the shape characteristics of objects, allowing the network to maintain shape similarity between the objects in the ground truth and the prediction maps. Different than the previous methods, this loss function is able to adapt itself to give different degrees of importance to learning different levels of shape details during network training. The proposed work relies on quantifying shape (dis)similarity with a set of Fourier descriptors and penalizing this dissimilarity in a loss function, which we call \textit{FourierLoss}, in an adaptive manner. To this end, we first quantify the shape of an object with a function defined on the object's contours, expand this function in a Fourier series, and use the harmonic amplitudes of its Fourier coefficients as the object's Fourier descriptors. Each Fourier descriptor gives a different level of detail about this function and hence about the object's shape. We then define \textit{FourierLoss} as a weighted cross-entropy, where the weights are defined as the linear combination of the differences between the Fourier descriptors of the objects in the ground truth and the prediction maps. The coefficients in this linear combination determine the effects of the descriptors on the loss function, and thus, their selection will greatly affect the segmentation results. This work considers the selection of appropriate coefficients as part of the training procedure, and proposes to end-to-end learn them, simultaneously with the network weights by backpropagation. 

The contributions of this work are two folds: First, it quantifies the shape dissimilarity between the ground truth and the predicted objects using a set of Fourier descriptors, and proposes to use this dissimilarity as a penalty term in the loss function. In our previous work~\cite{cansizFourierNetShapePreservingNetwork2023}, we also used the Fourier descriptors to characterize the object shape. On the other hand, different than our current proposal, this previous work generated regression maps using the Fourier descriptors of the ground truth objects, defined their predictions as auxiliary tasks, and formulated segmentation as concurrent learning of these regression maps and the main task of pixel label prediction. It did not calculate the Fourier descriptors on the predicted objects, nor did it use the difference between these descriptors and those of the ground truth objects to define a weighted loss function, unlike the \textit{FourierLoss} definition. There also exist studies that defined the Fourier coefficients on the contour points to represent the object shapes and estimated them through an encoder network~\cite{jeon2022fcsn,riaz2021fouriernet}. These studies used the Fourier coefficients estimated by the network either for reconstructing the contour points~\cite{riaz2021fouriernet} or directly in a loss function to train this encoder network~\cite{jeon2022fcsn}. However, different than the proposed \textit{FourierLoss}, they did not involve any segmentation mask prediction during the training phase of this encoder network, and hence, they did not define a loss function to minimize the difference between the Fourier descriptors of the ground truth and the prediction maps. Moreover, these previous studies did not define an adaptive loss function with trainable parameters, as to be pointed out in the second contribution. Thus, all these previous studies~\cite{cansizFourierNetShapePreservingNetwork2023,jeon2022fcsn,riaz2021fouriernet} are different than our current proposal.

\begin{figure}
\centering
\small{
\begin{tabular}{@{~}c@{~}c@{~}c@{~}c@{~}}
\includegraphics[width =0.22\columnwidth]{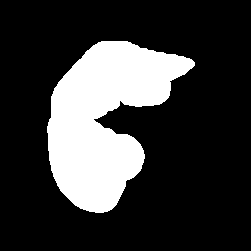} &
\includegraphics[width =0.22\columnwidth]{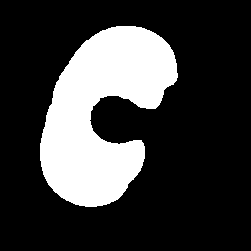} & 
\includegraphics[width =0.22\columnwidth]{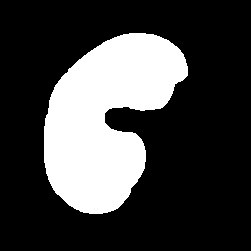} &
\includegraphics[width =0.22\columnwidth]{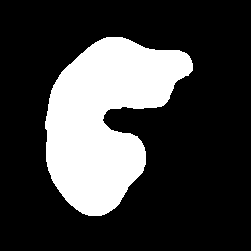} \\
(a) & (b) & (c) & (d) \\
\end{tabular}
}
\caption{(a) Ground truth object. Objects reconstructed from (b) the first, (c) the first two, and (d) the first eight Fourier coefficients of the function defined on the object's boundaries. In each figure, object boundaries are first reconstructed and the objects are obtained by filling the reconstructed boundaries.} 
\label{fig:fdmap}
\end{figure}

Second, this work introduces a loss function with trainable hyperparameters that control the importance of shape granularity in the training process; i.e., they control whether the network will focus on learning the general outline or detailed contour points. This is achieved thanks to the unique characteristics of a Fourier series, which allows a function to be reconstructed from its Fourier coefficients. The reconstructed function will approach to the original one as the number of the coefficients increases. In our case, the first Fourier coefficient gives the general outline (coarse details) of the object, whereas subsequent ones add more and more finer details to this outline (see Figure~\ref{fig:fdmap}). Since \textit{FourierLoss} penalizes the error made for each Fourier descriptor by a different amount, specified by the corresponding hyperparameter, it allows the network to pay different attention to learning different levels of shape detail. Furthermore, since these hyperparameters are end-to-end learned by backpropagation during training, the network can dynamically change its attention to emphasize different shape details in different training periods. As a result, \textit{FourierLoss} can balance coarse and fine shape details in network training without increasing model complexity, allowing the network to focus on more essential tasks early in training and learn more details later. Such kind of adaptive loss update mechanism has not been proposed before by the previous studies.

In summary, this work has two main contributions:
\begin{itemize}
\item It proposes a new shape-aware loss function based on quantifying shape dissimilarity with a set of Fourier descriptors and penalizing this shape dissimilarity between the ground truth and the predicted objects. 
\item It introduces an adaptive loss function with trainable hyperparameters that allows the network to dynamically change its attention to emphasize different shape details in different training periods. 
\end{itemize}

We tested our proposed \textit{FourierLoss} function in two different network architectures for the purpose of liver segmentation in CT images. Our experiments on 2879 images of 93 subjects revealed that this adaptive shape-aware loss function with trainable hyperparameters led to statistically significantly better results compared to its counterparts. This improvement was attributed to the networks' ability to learn the context better, which in turn made them more robust to local variations. 

\section{Related Work}

Segmentation models proposed in the last decade typically included encoder-decoder networks that predicted the output for each pixel separately and were trained by minimizing a loss function that was an aggregate of losses defined for these pixels independently. As a drawback of this design, these networks mostly relied on capturing local patterns and may fail to utilize the global context of the entire image to make their predictions. To mitigate this problem by modeling the global context, more recent studies proposed to make use of shape-related information. Previous studies integrated shape-related information into their designs either by proposing loss functions or by utilizing shape priors in their network architectures. For example, in~\cite{alarifShapeAwareDeepConvolutional2018}, it was proposed to calculate the average Euclidean distance between the ground truth and the predicted objects and use this distance as a penalty term in the cross-entropy loss for false negative and false positive pixels. In another study \cite{chen2019learning}, an energy term built on the contour length and the region information was minimized as a loss function. In \cite{zhang2022shape}, the authors penalized the inconsistency between the predictions of original and cut-out inputs in the loss function to exploit shape information. Predefined filters \cite{tofighi2019prior} and pixel-wise probability of each class based on the ground-truth labels \cite{zotti2018convolutional} were proposed to be employed as shape priors in the network architectures. However, none of these studies quantified varying shape characteristics and considered the harmonious balance between them during network training. 

Alternatively, instead of predicting the entire segmentation masks, some other studies quantified the shape of these masks with parametric equations and predicted their parameters using a neural network. In~\cite{liu2020abcnet}, a neural network was trained to predict the parameterization of the Bezier curves, which fitted to the boundary curves for scene text detection. Similarly, in~\cite{chen2023bezierseg}, the parametric representations of the Bezier curves encompassing the object boundary in medical images were learned. Apart from them, in~\cite{xie2020polarmask}, the polar coordinates of the sampled contour points were predicted via a neural network for instance segmentation. In~\cite{xu2019explicit}, the authors parameterized the contour points of objects in polar coordinates and trained a neural network to estimate the corresponding coefficients as shape descriptors.

The distance transforms were also employed to model the shape information in the existing convolutional neural network pipelines. One common approach was to calculate a distance transform map on the segmentation mask and learn its prediction by a single-task regression network~\cite{sironi2014multiscale,xue2020shape}. Alternatively, this prediction was defined as an auxiliary task and learned together with the main task of segmentation in a multi-task learning framework~\cite{murugesan2019psi, wang2020deep}. In this approach, the network may fail to capture the overall shape characteristics of the foreground regions effectively since a distance map covers all pixels in the image, and estimating it as a regression task necessitates focusing on also background pixels along with the foreground ones. It was also proposed to make use of distance transforms in the loss function definition. For example, in~\cite{karimi2019reducing}, the authors utilized the distance transforms of the ground truth objects to approximate the one-sided Hausdorff distance and used it as a penalty term in the loss function. Since the Hausdorff distance is based on measuring the maximum shape deviation, its use may not help accurately capture the distances between the ground truth and the predicted object boundaries in their entirety, especially when the distribution of these distances was skewed. 

There also exist previous studies that used Fourier coefficients to represent the object shapes. In our earlier work~\cite{cansizFourierNetShapePreservingNetwork2023}, we defined a map of the Fourier descriptors by iteratively calculating them on the object contours and formulated the segmentation task as a consecutive learning of this map and the segmentation mask in a cascaded network design. In~\cite{jeon2022fcsn,riaz2021fouriernet}, the Fourier coefficients of the contour points were defined, and their estimations were learned by training an encoder network. In~\cite{jeon2022fcsn}, the Fourier descriptors of the sampled boundary points of the ground truth objects were calculated in a complex domain, and a neural network was trained to predict these descriptors. In this training, the differences between the actual and estimated Fourier descriptors were used in the loss function definition. For a test image, the Fourier descriptors were predicted by the trained network and the segmentation mask was generated by applying the inverse Fourier transform on the predicted descriptors. Different than our proposal, in this work, the network training did not involve any segmentation map prediction and did not use an adaptive loss function with trainable hyperparameters. Likewise, in~\cite{riaz2021fouriernet}, the Fourier coefficients of the contour points were predicted by a neural network. Then, the loss function was defined on the contour points reconstructed from the predicted Fourier coefficients. On the other hand, in contrast to our current proposal, this previous work did not calculate the Fourier coefficients of the ground truth mask during training, and hence, did not employ these coefficients in its loss function.

\begin{figure*}
\centering
\includegraphics[height=6.4cm]{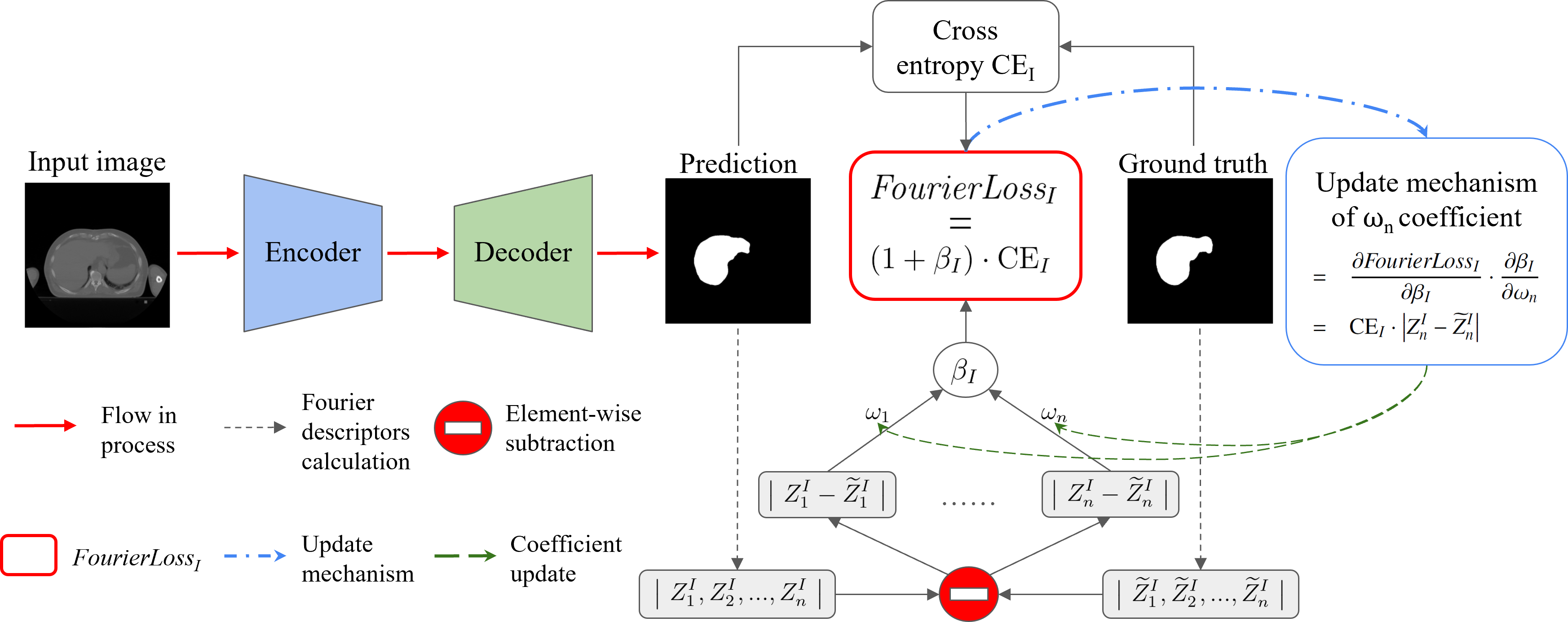}
\caption{Illustration of the learning procedure with the proposed \textit{FourierLoss} function and the loss update mechanism.} 
\label{fig:overview}
\end{figure*}

\ul{\textit{Liver segmentation in CT images}}: Atlas-based techniques, with a limited set of features, were previously used for liver segmentation~\cite{li2010fully, xu2015efficient}. However, they were 
prone to over/under segment the regions of interest due to lacking generalizations for the variations of the liver shape across axial slices and subjects. Since the UNet model~\cite{ronneberger2015u} had gained popularity for medical image segmentation, more recent studies also used U-shaped networks with several modifications for liver segmentation. These modifications included the uses of residual blocks \cite{manjunath2022automatic,sabir2022segmentation} and spatial channel-wise convolutions~\cite{chen2019channel} in the network design. Cascaded networks were also proposed to achieve liver segmentation and related tasks, such as lesion segmentation and liver boundary refinement, in line~\cite{araujo2022liver,christ2017automatic,tang2018dsl}. Although these previous studies led to promising results, liver segmentation in CT images still remains a challenging problem due to the close pixel intensities of the liver and its surroundings, and also due to the variations in the liver shapes across the axial slices of the same subject as well as the slices of different subjects. None of the previous studies employed a loss function that penalized the shape dissimilarity based on the Fourier descriptors and in an adaptive manner. Thus, different than these studies, our work proposes a new approach for liver segmentation in CT images.

\section{Methodology}

This work relies on calculating Fourier descriptors on the ground truth and the predicted objects, penalizing the dissimilarity between these descriptors in its
\textit{FourierLoss} definition, and devising an update mechanism to dynamically change this loss function throughout network training. The details of the Fourier descriptor calculation, the \textit{FourierLoss} definition, and the loss update mechanism are explained below. The overview of the proposed learning framework is illustrated in Figure~\ref{fig:overview}.

\subsection{Fourier Descriptor Calculation}

\begin{figure}
\centering
\includegraphics[height=3.4cm]{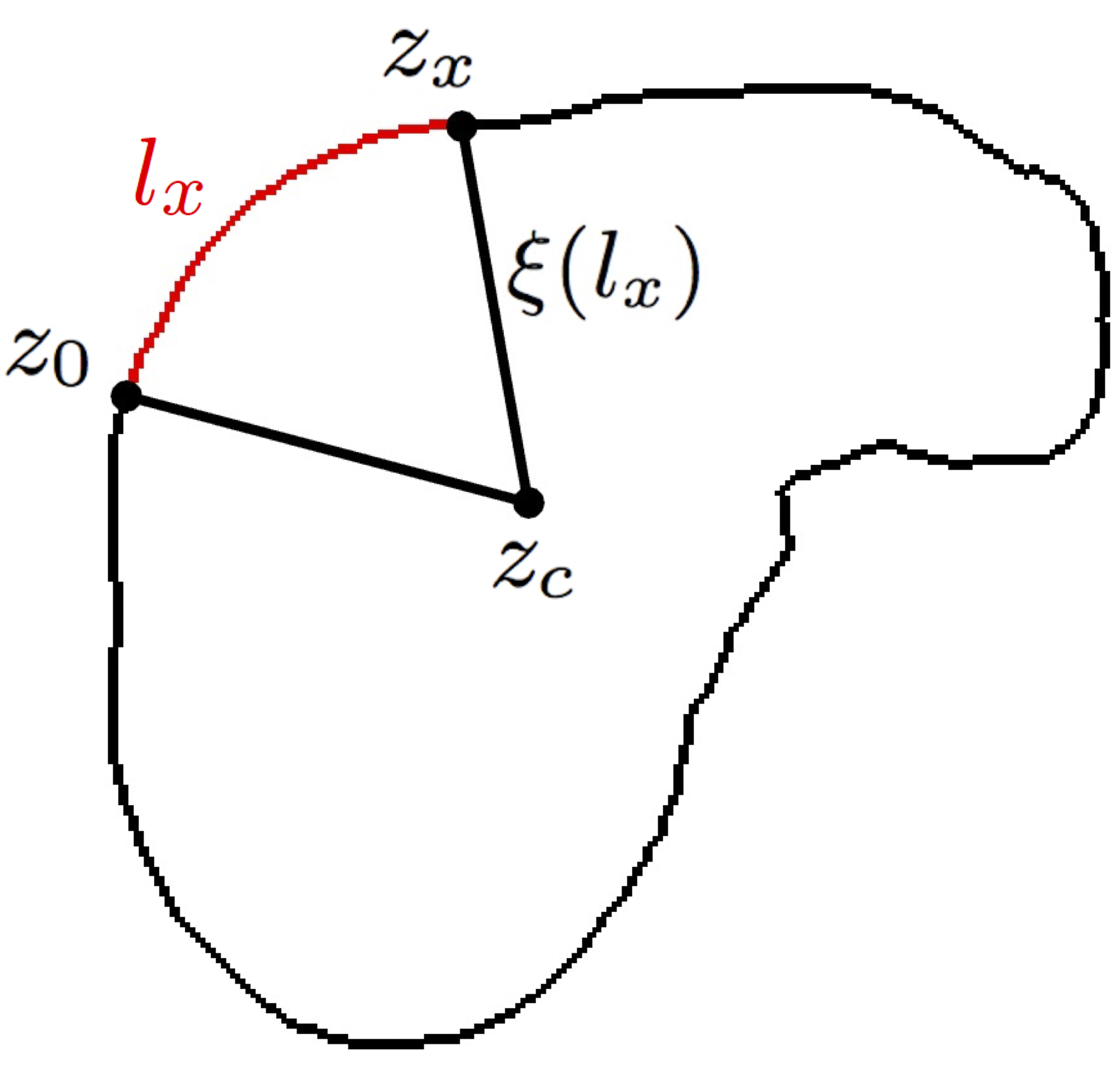}
\caption{The distance-to-center function $\xi(l_x)$ defined for $l_x \in [0,L]$, which is the arc length from the starting point $z_0$ to a contour point $z_x$.}
\label{fig:distance-to-center}
\end{figure}

The shape of the ground truth or the predicted object is quantified by the Fourier descriptors of the distance-to-center function defined along the object's contour points. Let $\gamma$ be the object's contour with length $L$, $z_0$ its starting point, $z_c$ its centroid, $z_x$ an arbitrary point on $\gamma$, and $l_x \in [0, L]$ an arc length from $z_0$ to $z_x$. The distance-to-center function $\xi(l_x)$ is defined on the arc length $l_x$ as the distance from the corresponding point $z_x$ to the centroid $z_c$ (see Figure~\ref{fig:distance-to-center}). We expand this function in a Fourier series as
\begin{align}
\label{elx}
\xi(l_x) = a_0 + \sum_{n=1}^{\infty} \left[a_n \cos\left(\frac{2\pi n l_x}{L} \right) + b_n \sin\left(\frac{2\pi n l_x}{L}\right)\right]
\end{align}
where
\begin{eqnarray}
\label{ak}
a_n &=& \frac{2}{L} \int_{0}^{L} \xi(l_x) \cos\left(\frac{2\pi n l_x}{L}\right) dl_x\\ 
\label{bk}
b_n &=& \frac{2}{L} \int_{0}^{L} \xi(l_x) \sin\left(\frac{2\pi n l_x}{L}\right) dl_x
\end{eqnarray}

Since the contour of an object in a digital image is not continuous but contains a finite number of discrete points (pixels), we calculate the Fourier coefficients $a_n$ and $b_n$ as given in Equations~\ref{eqn:fd-a} and~\ref{eqn:fd-b}, respectively, assuming an arc interpolation between the consecutive contour pixels. We refer the reader to our previous work~\cite{cansizFourierNetShapePreservingNetwork2023} for the derivation of the equations under this assumption. Here we would like to once again note that our previous work used this Fourier descriptor calculation to generate regression maps in a cascaded network design but did not use the Fourier descriptors in defining a shape-aware loss function.
\begin{eqnarray}
a_n &=& \frac{1}{\pi n} \sum_{t=1}^{T}   \Delta \xi_t~\sin\left(\frac{2\pi n l_{t}}{L}\right) \label{eqn:fd-a}\\
b_n &=& -\frac{1}{\pi n} \sum_{t=1}^{T}   \Delta \xi_t~\cos\left(\frac{2\pi n l_{t}}{L}\right) \label{eqn:fd-b}
\end{eqnarray}
In these equations, $\Delta \xi_t=\xi(l_{t-1})-\xi(l_t)$. For the $n$-th Fourier coefficients $(a_n, b_n)$, the harmonic amplitude is denoted as $Z_n = \sqrt{{a_n}^2+{b_n}^2}$. The first $N$ harmonic amplitudes $[Z_1, Z_2, ..., Z_N]$ are then used as the Fourier descriptors of the object. 

\subsection{FourierLoss Definition}
Let $I$ be a training image and $q \in I$ be a pixel. The binary cross-entropy loss $\texttt{CE}_I$ of this image is defined as
\begin{equation}
\text{CE}_I = \sum_{q\in I}- p_q~\log \hat{p}_q - (1-p_q)~\log(1-\hat{p}_q)
\label{eqn:bce}
\end{equation}
where $p_q$ is the class label of the pixel $q$ in the ground truth map $G_I$, and $\hat{p}_q$ is its posterior in the predicted segmentation map $S_I$. Here $p_q = 1$ for foreground pixels and $p_q = 0$ otherwise.

The \textit{FourierLoss} function penalizes the shape dissimilarity between the ground truth and the predicted objects. To this end, it is defined as a weighted cross-entropy, where the weights are added as regularization terms to emphasize the ground truth objects that show shape inconsistency with their corresponding predicted objects, and to learn how to correct this inconsistency. For the image $I$, it is defined as
\begin{eqnarray}
\textit{FourierLoss}_I &=& (1 + \beta_I) \cdot \text{CE}_I \\
\beta_I &=& \sum_{o \in G_I} \beta(~o,~\sigma(o)~) \label{eqn:beta}
\end{eqnarray}
where the term $\beta(~o,~\sigma(o)~)$ denotes the shape dissimilarity between a ground truth object $o \in G_I$ and its matching predicted object $\sigma(o) \in S_I$. Here each ground truth object $o$ must be matched with a predicted object $\sigma(o)$ after a forward pass based on the predicted segmentation map $S_I$. In our experiments, first, we only consider the largest object $O_I$ in the ground truth map $G_I$ and match it with the largest predicted object $\sigma(O_I) \in S_I$ for the sake of simplicity, since almost all CT images consist of a single liver region. Thus, we simplify Equation~\ref{eqn:beta} as
\begin{equation}
    \beta_I = \beta(~O_I,~\sigma(O_I)~)
\end{equation}

Here it is worth noting that the formulation in Equation~\ref{eqn:beta} allows to define $B_I$ over all ground truth objects for multi-instance object segmentation tasks. In that case, each ground truth object can be matched with its maximally overlapping predicted object. In our application, this is not necessary for many images since there are usually no multiple liver regions to be segmented in a typical CT image. Thus, we first consider the largest ground truth object only for the matching in the first network we use, which is a UNet based model. Nevertheless, to explore the possibility of using more than one match, in another network design, we include every ground truth object $o \in G_I$ in the calculation of $\beta_I$, provided that the intersection-over-union score between $o$ and a predicted object $\sigma(o)$ is greater than 0.5. Note that all equations given in the remaining part of this section are derived for the former case, in which only the shape dissimilarity between the largest objects, $\beta(~O_I,~\sigma(O_I)~)$, is used. For the latter case, the equations need to be modified to include the summation over all of the matching objects $\beta(~o,~\sigma(o)~)$. 

The shape dissimilarity term $\beta(~O_I,~\sigma(O_I)~)$ is defined as a linear combination of the differences between the first $N$ Fourier descriptors $Z_n^I$ of the largest ground truth object $O_I$ and the descriptors $\widetilde{Z}_n^I$ of the corresponding predicted object $\sigma(O_I)$. Thus, for the image $I$, the weight term $B_I$ is expressed as
\begin{equation}
    \beta_I = \beta(~O_I,~\sigma(O_I)~) = \sum_{n=1}^N \omega_n \cdot \big\lvert Z_n^I - \widetilde{Z}_n^I \big\rvert
\end{equation}
where the coefficient $\omega_n$ determines the effect of the $n$-th descriptor to the loss function. We consider these coefficients as the network hyperparameters, and end-to-end learn them simultaneously with the network weights by backpropagation. At each epoch of the backpropagation algorithm, the forward pass estimates the segmentation maps $S_I$ using the current network weights, calculates the Fourier coefficients $\widetilde{Z}_n^I$ of the largest object in $S_I$, and updates $\textit{FourierLoss}_I$ using the current $\omega_n$ coefficients. Then, the backward pass updates the network weights together with the coefficients $\omega_n$ by differentiating the updated loss function $\textit{FourierLoss}_I$. For each image, the update amount in $\omega_n$ is calculated as
\begin{eqnarray}
\frac{\partial \textit{FourierLoss}_I}{\partial \omega_n} &=& \frac{\partial \textit{FourierLoss}_I}{\partial \beta_I}  \cdot \frac{\partial \beta_I}{\partial \omega_n} \\
&=& \text{CE}_I  \cdot \big\lvert Z_n^I - \widetilde{Z}_n^I \big\rvert
\label{eqn:derivation}
\end{eqnarray}

In this equation, the first term $\text{CE}_I$ is the cross-entropy, which is the same for all Fourier descriptors of a given image. The second term $\big\lvert Z_n^I - \widetilde{Z}_n^I \big\rvert$ is different for each Fourier descriptor. It determines the required increase in the importance of learning that particular descriptor in the next epoch, and thus, the importance of learning a level of the shape detail to which that Fourier descriptor corresponds. For example, if this term is relatively large for the first Fourier descriptor, the network will pay more attention to correcting the general outline (coarse details) of the predicted object in the next epoch. On the other hand, if it is smaller than those of the latter descriptors, the network will attend finer details of this outline. 

This update mechanism gives \textit{FourierLoss} the ability of being adaptive to dynamically shift the network's attention to emphasize different shape details in different training epochs. There are two points to consider with this mechanism. First, since $B_I$ is a linear combination, the ratio between the updated $\omega_n$ coefficients, rather than their absolute values, determines the relative impact of each Fourier descriptor on the loss function, which will affect the updates of the network weights in a given epoch. Second, in each iteration, we increase the $\omega_n$ coefficient by adding the term given in Equation~\ref{eqn:derivation} multiplied with a learning rate. The motivation here is our intention to have the Fourier coefficients (shape awareness) increasingly impact the loss function in later epochs, especially before the training stops. To allow the network to achieve a better balance between learning network weights and shape awareness, one has the flexibility to select this learning rate different than the one used for network weight updates.

\subsection{Network Architectures and Training}

 \begin{figure*}[ht]
    \centering
    {\includegraphics[width=0.85\textwidth]{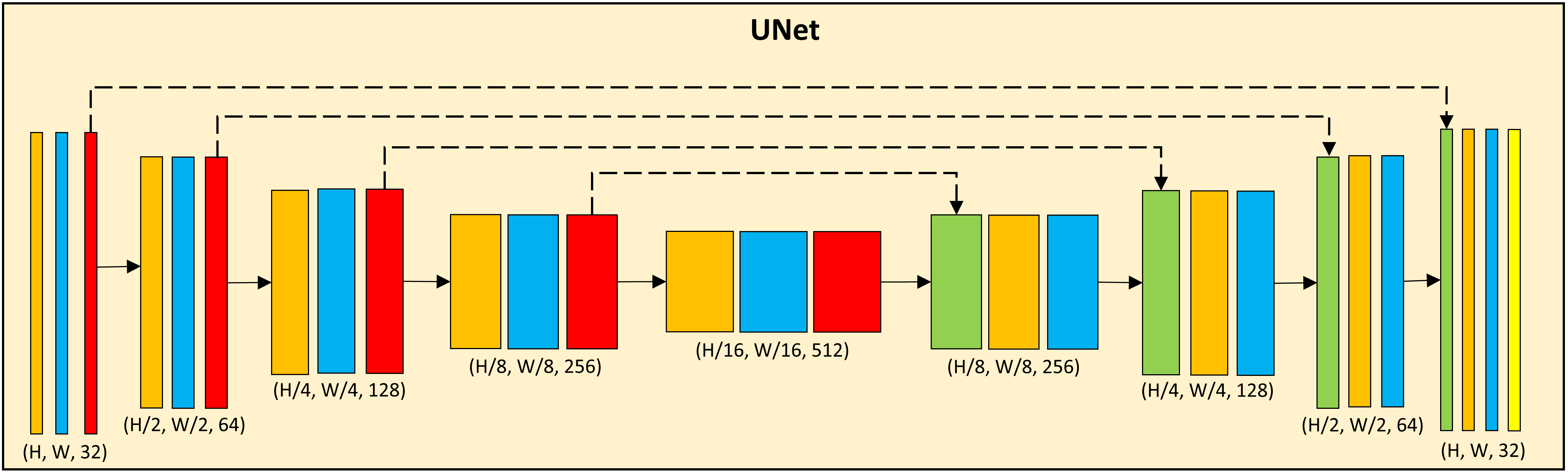}}
    \\
    {\includegraphics[width=0.85\textwidth]{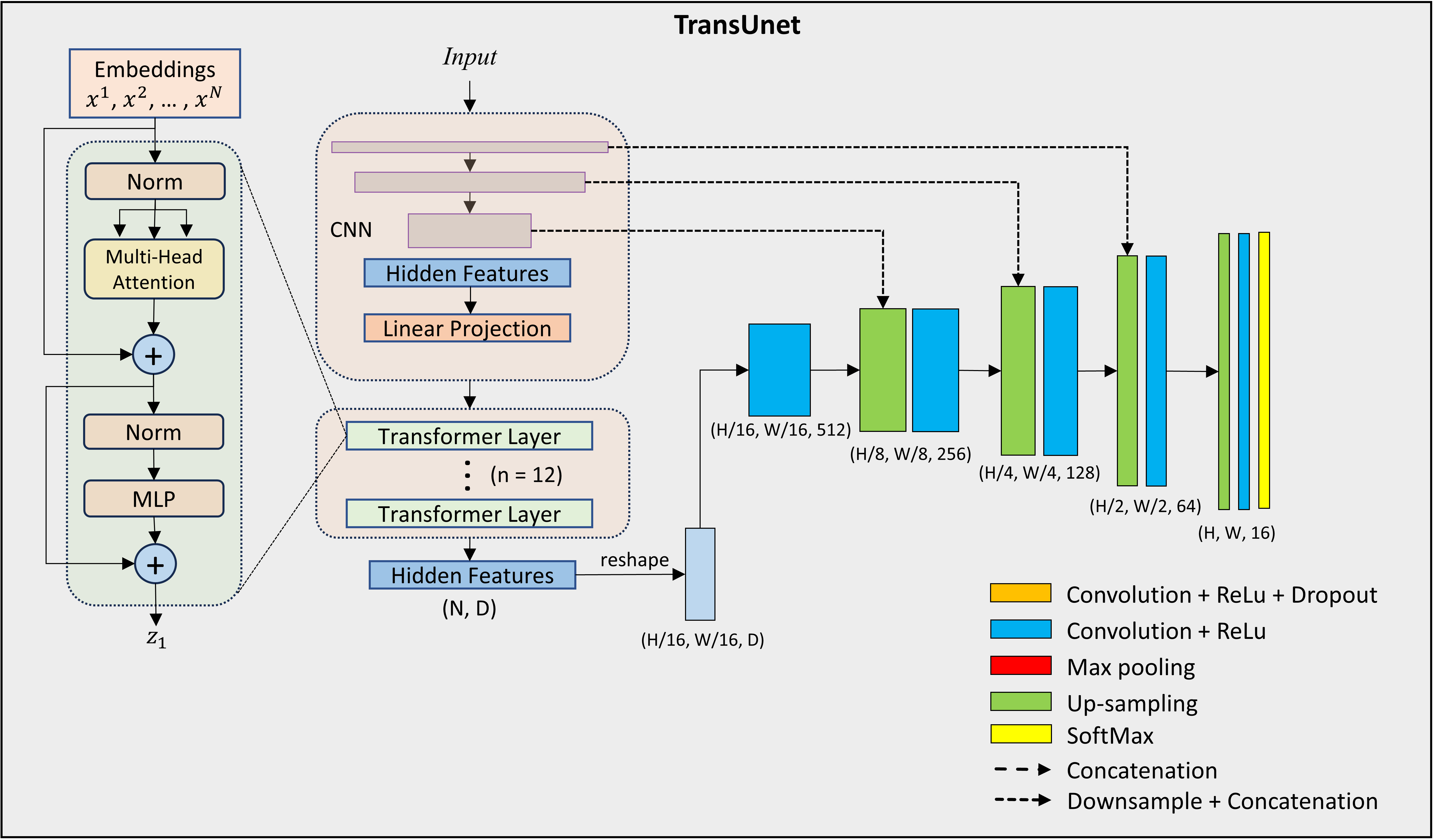}} 
    \caption{Pictorial overview of the UNet and TransUnet architectures used in the experiments. Here $H$ and $W$ denote the height and width of the input image, respectively, and $N$ and $D$ denote the number of patches and embedding dimension in transformer layers, respectively.}
\label{fig:networks}
\end{figure*}

\textit{FourierLoss} offers a generic loss function that can be used with any network architecture. In this work, we use two architectures to observe the effect and contribution of the novel \textit{FourierLoss} modification in the networks with different complexities. The first one is the UNet~\cite{ronneberger2015u}, which is known as one of the fundamental architectures for medical image segmentation. We select this architecture for our experiments due to its simplicity together with its high performance and prevalence on medical image segmentation problems. We use the UNet architecture illustrated in Figure~\ref{fig:networks} with $3\times3$ kernels in the convolution layers and $2\times2$ kernels in the max pooling/upsampling layers. A dropout layer with 0.2 dropout factor is added after the first convolution layer of each block to avoid overfitting. The number of feature maps in the first encoder block is selected as 32. Similar to the original design~\cite{ronneberger2015u}, the number of feature maps is doubled at the end of each encoder block and halved at the end of each decoder block. In the training of this UNet model, the Adam optimizer with a learning rate of \num{5.0e-4} is used for the network weights, and the batch size is set to 8. The learning rate for the Fourier coefficients $\omega_n$ is selected as \num{1.0e-3}. The training continues for a maximum of 500 epochs with an early stopping option. The training stops if there is no improvement in the calculated \textit{FourierLoss} on the validation images for the last 20 epochs. Note that in our experiments, all models stopped before reaching this maximum number thanks to early stopping.

The second architecture is TransUnet~\cite{chen2021transunet}, a more recent encoder-decoder design that uses an attention mechanism with a transformer block. The decoder path of the TransUnet model is similar to that of the UNet model. However, 12 consecutive transformer layers are embedded in the network's encoder, making it more complex and advanced than the baseline UNet. We select this architecture for our experiments as it was shown to yield high performance also on medical image datasets. Its encoder includes ResNet-50~\cite{he2016deep}, which was pretrained on the ImageNet dataset~\cite{deng2009imagenet}, to obtain high-level feature representations from inputs. Patch embeddings for the transformer layers are mapped from hidden features using a trainable linear projection. The decoder path upsamples the latent representations coming out of the transformer layers to the full spatial resolution. Similarly, the decoder path uses $3\times3$ and $2\times2$ kernels for the convolution and upsampling layers, respectively. The illustration of the TransUnet architecture is also depicted in Figure~\ref{fig:networks}. In its training, batch size is decreased to 4, the learning rates of the network weights and coefficients $\omega_n$ are increased to \num{1.0e-2}, and the maximum of 50 epochs is used by considering the hardware limitations and the expected performance of the TransUnet model. 

Both networks are implemented using the PyTorch framework. The training of both of the networks starts with a warm-up period for five epochs where only the standard cross-entropy loss is minimized. We use this warm-up period since the first epochs may produce inaccurate predictions, for which the Fourier descriptors are irrelevant to the actual ones, and this may cause divergence or fluctuations in the network training. 

\section{Experiments}

\subsection{Dataset}
The proposed \textit{FourierLoss} was tested on a dataset that contains 2879 CT images of 93 subjects, which underwent radiotherapy +/-chemotherapy with lung cancer at the Radiation Oncology Department of Eskisehir Osmangazi University Faculty of Medicine between 2014 and 2022. The subjects were immobilized in a supine position using T-bar/Wingboard with their hands above their head and 3-5 mm slice thickness of images between the cricoid cartilage and the upper limit of the L2 vertebra were performed with the Siemens Somatom Definition AS\textregistered~CT device. The input image resolution was 512$\times$512 pixels. Livers were contoured by two radiation oncologists with at least 10 years of experience. The data collection was conducted in accordance with the tenets of the Declaration of Helsinki and was approved by Eskisehir Osmangazi University Institutional Review Board (Protocol number: E-25403353-050.99-338117).

We randomly split the 93 subjects into the training, validation, and test sets. The training set contains 1510 CT images of 50 patients, the validation set contains 455 CT images of 15 patients, and the test set contains 914 CT images of the remaining 28 patients. For the UNet architecture, the training images are used to update the weights in backpropagation and validation images for early stopping. However, since we do not use early stopping to train the TransUnet architecture, we use the training and validation images together in backpropagation.

\subsection{Evaluation}

Liver segmentations were evaluated visually and quantitatively on the test set images. Comparing the ground truth and the predicted segmentation maps, the number of true positive (TP), false positive (FP), and false negative (FN) pixels were found. Then, for each image $I$, the pixel-wise precision, recall, intersection over union (IoU), and f-score were calculated. 
\begin{eqnarray}
\text{precision} &=& \text{TP}~/~(\text{TP}~+~\text{FP})\\
\text{recall} &=& \text{TP}~/~(\text{TP}~+~\text{FN})\\
\text{IoU} &=& \text{TP}~/~(\text{TP}~+~\text{FP}~+~\text{FN})\\
\text{f-score} &=& 2\cdot \frac{\text{precision} \cdot \text{recall}}{\text{precision} + \text{recall}}
\end{eqnarray}
Additionally, for the image $I$, the Hausdorff distance was computed between the  ground truth $G_I$ and the predicted segmentation $S_I$ maps. For that, for each foreground pixel $q \in G_I$, the distance from $q$ to the closest foreground pixel in $S_I$ was calculated. Likewise, for each foreground pixel $r \in S_I$, the distance from $r$ to the closest foreground pixel in $G_I$ was calculated. The maximum of these distances was defined as the Hausdorff distance. All these performance metrics were calculated for each CT image separately, and then the averages over the test set images were used for quantitative evaluation. Higher precision, recall, IoU, and f-score values and lower Hausdorff distances indicate more accurate algorithms.

\subsection{Comparisons}
\label{sec:comparison}

As aforementioned, we tested the proposed \textit{FourierLoss} function on two architectures: UNet and TransUnet. For the UNet architecture, we used five algorithms for comparison and ablation studies. The first one is the \textit{BaselineMethod} that had the same UNet~\cite{ronneberger2015u} architecture given in Figure~\ref{fig:networks} but used the standard cross-entropy loss instead of our proposed loss function. Its architecture and training settings were exactly the same with ours. We included this method into our comparisons to observe the effects of using a shape-aware loss function in network training. The next two algorithms, namely \textit{HausdorffDistanceMethod}~\cite{karimi2019reducing} and \textit{ActiveContourMethod}~\cite{chen2019learning}, were the state-of-the-art algorithms that proposed a shape-aware loss function for network training. For these two methods, we also used the same network architecture, given in Figure \ref{fig:networks}, but trained the network with their own loss functions. 

The \textit{HausdorffDistanceMethod} proposed a loss function that multiplied the mean squared error loss between the ground truth and the predicted probability maps by a Hausdorff distance term defined as $\Big(d(G_I)^\alpha + d(S_I)^\alpha\Big)$. In this term, $d(G_I)^\alpha$ and $d(S_I)^\alpha$ were considered as the Hausdorff distance maps that were proposed to be approximated by calculating the distance transforms on the ground truth $G_I$ and the predicted segmentation mask $S_I$, respectively~\cite{karimi2019reducing}. Note that since the computation of $d(S_I)^\alpha$ was required for all predicted images in each epoch, this previous study~\cite{karimi2019reducing} only considered $d(G_I)^\alpha$ as a one-sided Hausdorff distance term to reduce the computational cost. The parameter $\alpha$ determined how strongly large distances were penalized. In our experiments, we empirically selected $\alpha = 0.2$ for the UNet architecture and $\alpha = 0.5$ for the TransUnet architecture. The \textit{ActiveContourMethod} developed a loss function for shape preservation, inspiring by the general idea of active contour models. For that, it included the terms that were based on the segmentation contour length and the area inside and outside the ground truth and the segmentation masks into its loss function~\cite{chen2019learning}. We used these two methods, which also employed shape-preserving loss functions, in our comparisons in order to understand the effectiveness of our approach of quantifying shape dissimilarity with Fourier descriptors and penalizing this dissimilarity in a loss function. 

The fourth algorithm, \textit{CascadedFourierNet}, was our previously proposed model that designed a cascaded network to employ the Fourier descriptors for shape preservation~\cite{cansizFourierNetShapePreservingNetwork2023}. Although the descriptors were calculated in the same way, this previous work did not define any loss function on these descriptors, but used them to generate regression maps. It learned these regression maps in the first stage of its cascaded design, and then estimated the segmentation mask in its second stage from the input image along with the predicted regression maps. In other words, this comparison method employed the Fourier descriptors by making changes in the network architecture instead of defining a loss function. The last comparison method, \textit{NonAdaptiveFourierLoss}, was for an ablation study to understand the effects of using our proposed adaptive loss update mechanism with trainable hyperparameters. Its design and loss function definition were exactly the same with \textit{FourierLoss} except that $\omega_n$ values were initialized at the beginning and kept constant during the entire network training. In other words, its loss used the Fourier descriptors but did not have trainable hyperparameters. 

\captionsetup[table]{justification=centering, skip=2pt}

\begin{table*}[!ht]
\centering
\caption{Test set results obtained by our model and the comparison methods when the UNet architecture is used. Averages and standard deviations across five runs. Significantly best metrics ($p < 0.05$) are indicated in bold. For a selected metric, there is no statistically significant difference between the values that are all in bold.}
  \label{table:Unet-results}
  \begin{tabular}{lccccc}
\toprule
& Precision & Recall & F-score & IoU & Hausdorff d. \\ \hline  
BaselineMethod~\cite{ronneberger2015u} & 89.70~$\pm$~1.45 & 90.04~$\pm$~2.57 & 88.88~$\pm$~1.84 & 83.06~$\pm$~2.18 & 4.68$~\pm$~0.11 \\
HaussdorffDistanceMethod~\cite{karimi2019reducing} & \textbf{91.45~$\pm$~1.27} & 89.08~$\pm$~3.25 & 89.22~$\pm$~1.99 & 83.79~$\pm$~2.26 & 4.59~$\pm$~0.11 \\
ActiveContourMethod \cite{chen2019learning} & 90.64~$\pm$~0.56 & 89.96~$\pm$~2.59 & 89.26~$\pm$~2.14 & 83.86~$\pm$~2.09 & 4.60~$\pm$~0.15 \\
CascadedFourierNet~\cite{cansizFourierNetShapePreservingNetwork2023} & 90.05~$\pm$~0.77 & \textbf{93.07~$\pm$~1.82} & 90.68~$\pm$~0.75 & 84.77~$\pm$~0.88 & 4.68~$\pm$~0.06 \\
NonAdaptiveFourierLoss & 90.00~$\pm$~1.73 & 92.43~$\pm$~1.80 & 90.41~$\pm$~0.59 & 84.89~$\pm$~0.77 & 4.57~$\pm$~0.07 \\
\textit{FourierLoss} & 90.29~$\pm$~0.65 & \textbf{93.24~$\pm$~0.70} & \textbf{91.03~$\pm$~0.21} & \textbf{85.70~$\pm$~0.33} & \textbf{4.53~$\pm$~0.03} \\ 
\bottomrule
  \end{tabular}
\end{table*}

\begin{table*}[!ht]
\centering
\caption{Test set results obtained by our model and the comparison methods when the TransUnet architecture is used. Averages and standard deviations across five runs. Significantly best metrics ($p < 0.05$) are indicated in bold. For a selected metric, there is no statistically significant difference between the values that are all in bold.}
\label{table:Transunet-results}
  \begin{tabular}{lccccc}
    \toprule
& Precision & Recall & F-score & IoU & Hausdorff d. \\ \hline  
BaselineMethod~\cite{chen2021transunet} & 93.14~$\pm$~0.38 & 93.28~$\pm$~0.66 & 92.68~$\pm$~0.39 & 88.24~$\pm$~0.39 & 4.09~$\pm$~0.04 \\
HaussdorffDistanceMethod~\cite{karimi2019reducing} & 91.88~$\pm$~1.21 & 89.88~$\pm$~1.43 & 90.29~$\pm$~1.05 & 85.30~$\pm$~1.23 & 4.33~$\pm$~0.13 \\
ActiveContourMethod~\cite{chen2019learning} & 92.12~$\pm$~0.27 & \textbf{93.90~$\pm$~0.46} & 92.56~$\pm$~0.32 & 88.01~$\pm$~0.25 & 4.14~$\pm$~0.02 \\
NonAdaptiveFourierLoss & 92.95~$\pm$~0.76 & 93.45~$\pm$~0.56 & 92.60~$\pm$~0.34 & 88.20~$\pm$~0.41 & 4.12~$\pm$~0.11 \\
\textit{FourierLoss} & \textbf{93.45~$\pm$~0.28} & \textbf{93.94~$\pm$~0.47} & \textbf{93.20~$\pm$~0.18} & \textbf{88.77~$\pm$~0.21} & \textbf{4.06~$\pm$~0.02} \\ 
\bottomrule
\end{tabular}
\end{table*}

In our experiments, we trained our model as well as the comparison methods five times. Except \textit{CascadedFourierNet}, all methods used the same network architecture, and thus had the same learning capacity (i.e., the same number of network weights). To make fair comparisons between these methods, in each of the five training trials, the initial network weights for all methods were set to the same values using the same seed in the random number generation and processes of Pytorch. As a result, each method started with the same network weights and became different as a result of its loss function. Likewise, to make fair comparisons with the \textit{CascadedFourierNet} method, we selected the number of the Fourier descriptors it used as the same with the number that we used in this work.

For the TransUnet architecture, we used the same comparison methods except \textit{CascadedFourierNet} since this method was originally designed to consist of multiple UNet architectures in a cascaded network design. It would become too complicated with additional transformer layers and might have the risk of overfitting. Likewise, for the other comparison methods, we used the same network with the architecture given in Figure~\ref{fig:networks}, initialized the network weights using the same seed in the random number generation and processes of Pytorch, and trained each method five times. 

\begin{figure*}[ht]
    \begin{center}
    \begin{tabular}{@{~}c@{~}c@{~}c@{~}c@{~}c@{~}c@{~}c@{~}c@{~}}
    \includegraphics[width =0.225\columnwidth]{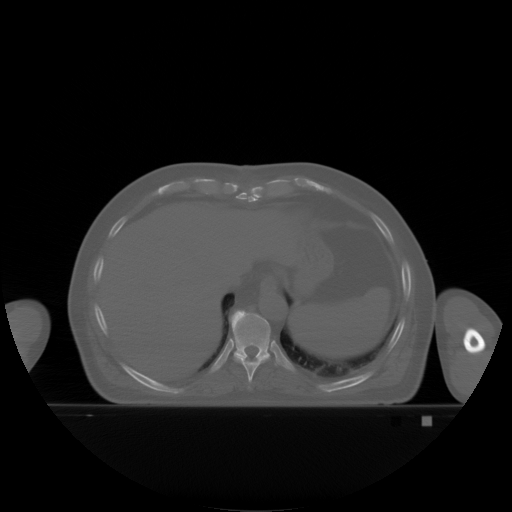} &
    \includegraphics[width =0.225\columnwidth]{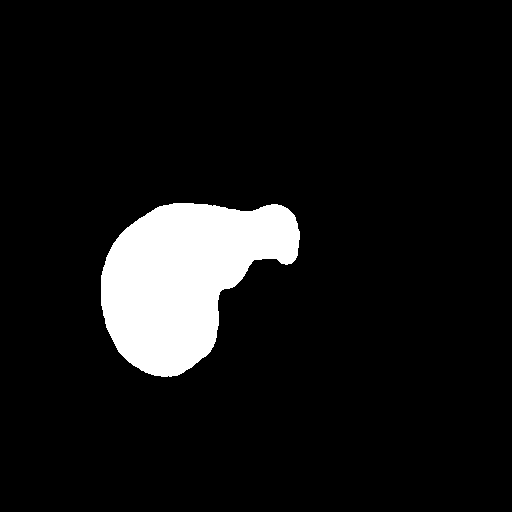} &
    \includegraphics[width =0.225\columnwidth]{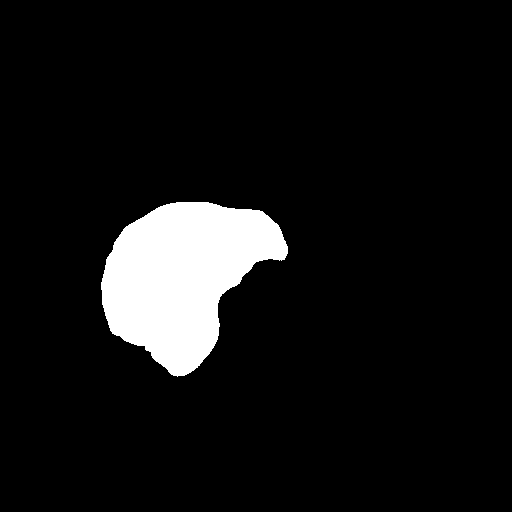} &
    \includegraphics[width =0.225\columnwidth]{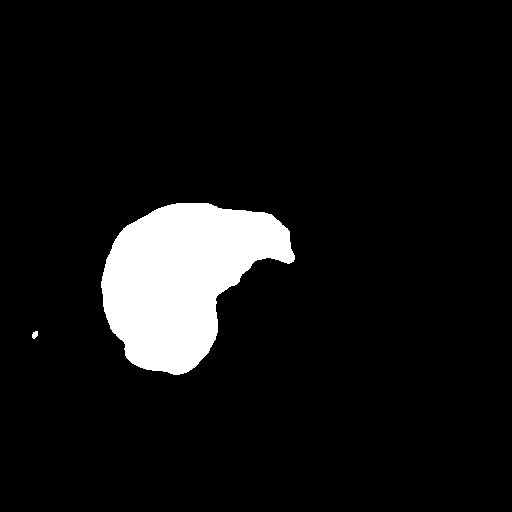} &
    \includegraphics[width =0.225\columnwidth]{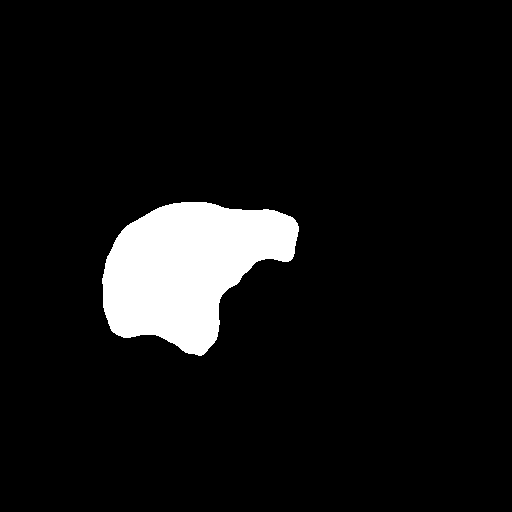} &
    \includegraphics[width =0.225\columnwidth]{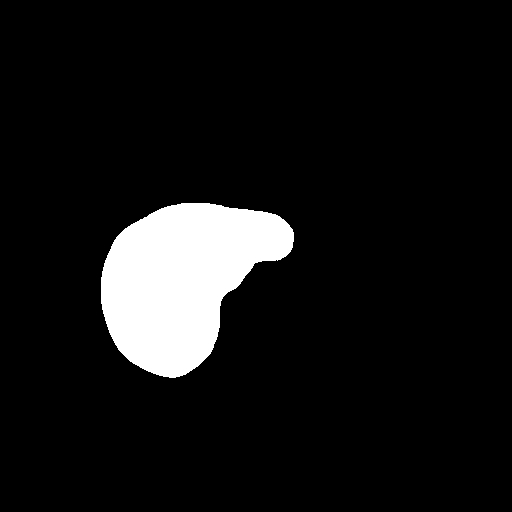} &
    \includegraphics[width =0.225\columnwidth]{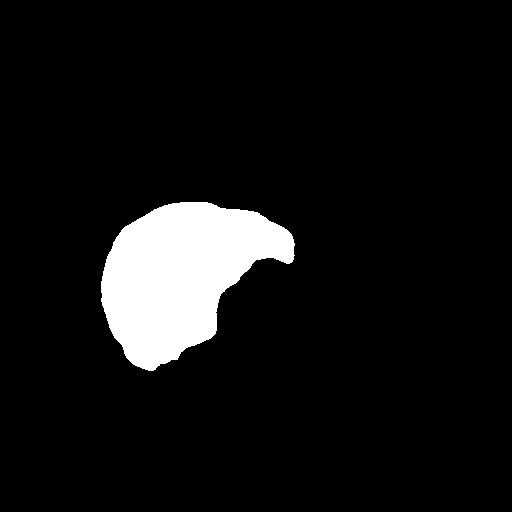} &
    \includegraphics[width =0.225\columnwidth]{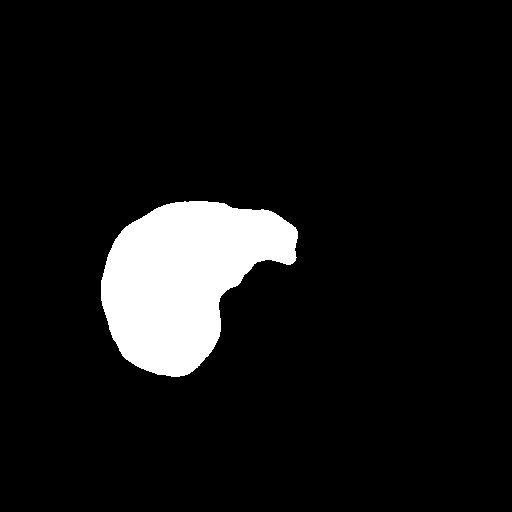} 
    \\
    \includegraphics[width =0.225\columnwidth]{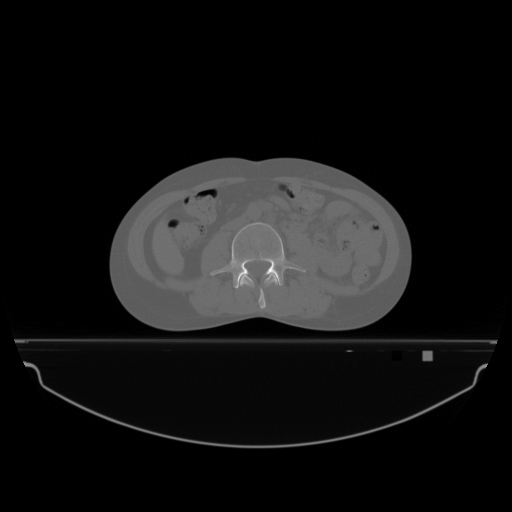} &
    \includegraphics[width =0.225\columnwidth]{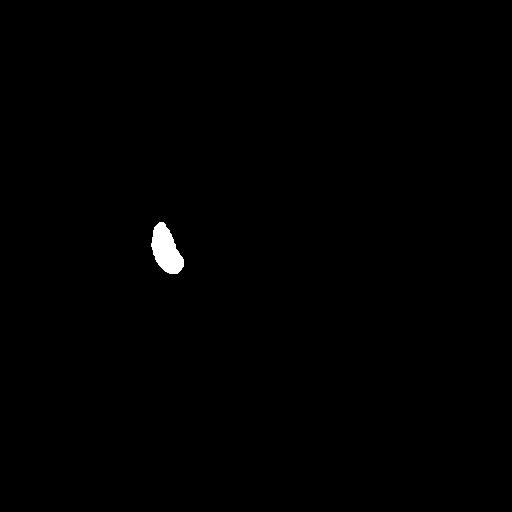} &
    \includegraphics[width =0.225\columnwidth]{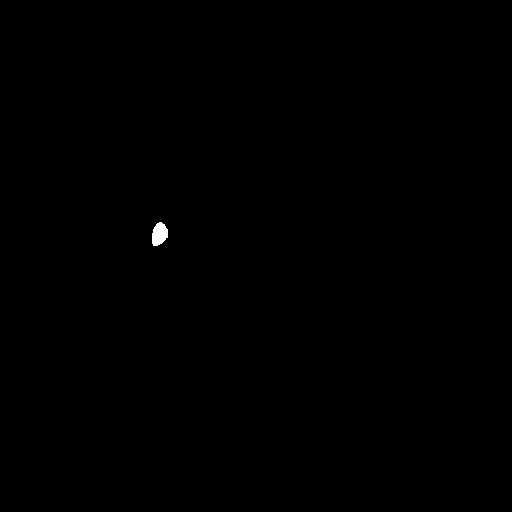} &
    \includegraphics[width =0.225\columnwidth]{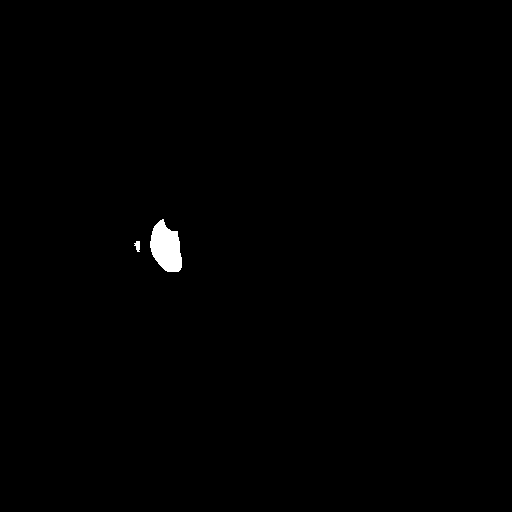} &
    \includegraphics[width =0.225\columnwidth]{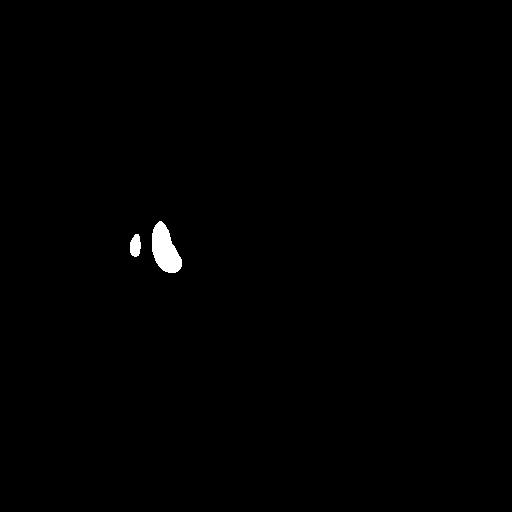}&
    \includegraphics[width =0.225\columnwidth]{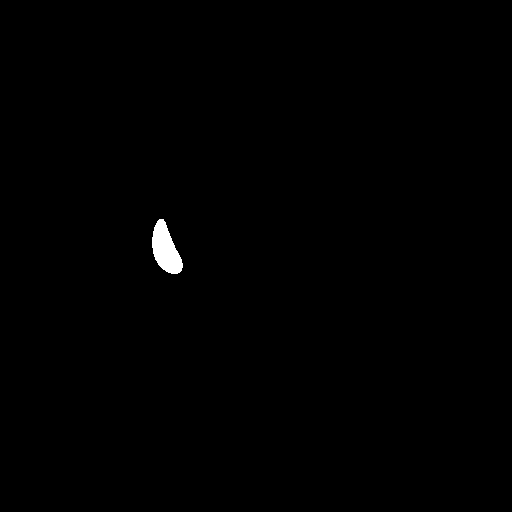}&
    \includegraphics[width =0.225\columnwidth]{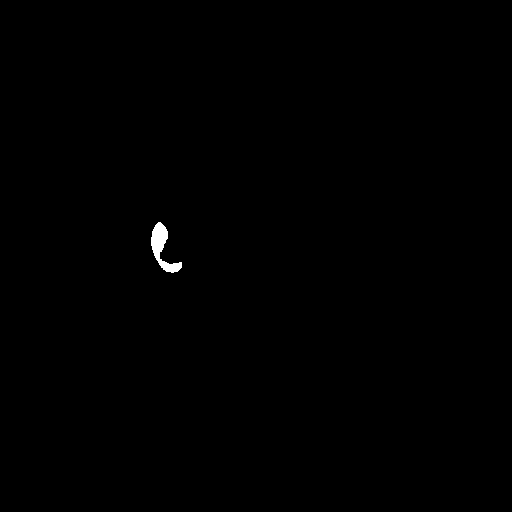} &
    \includegraphics[width =0.225\columnwidth]{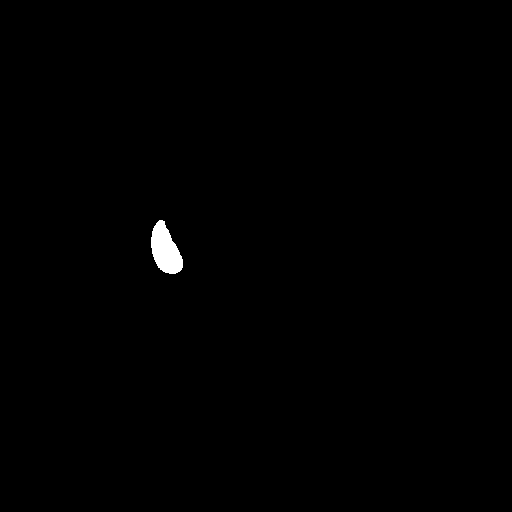}
    \\
    \includegraphics[width =0.225\columnwidth]{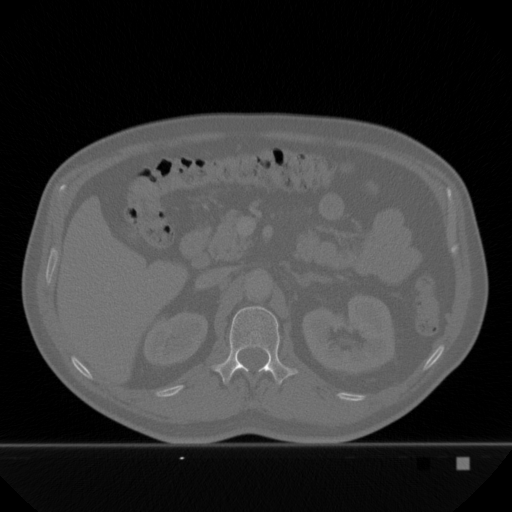} &
    \includegraphics[width =0.225\columnwidth]{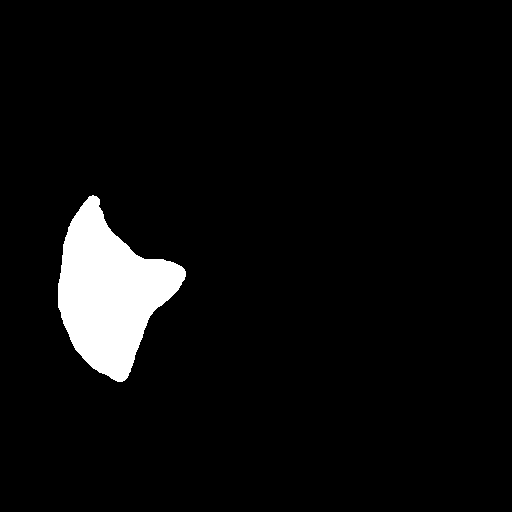} &
    \includegraphics[width =0.225\columnwidth]{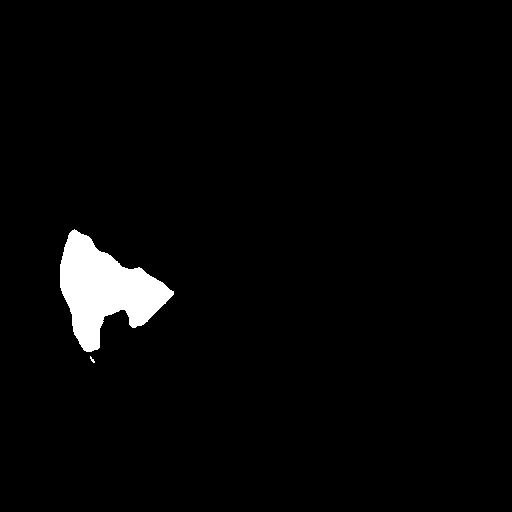} &
    \includegraphics[width =0.225\columnwidth]{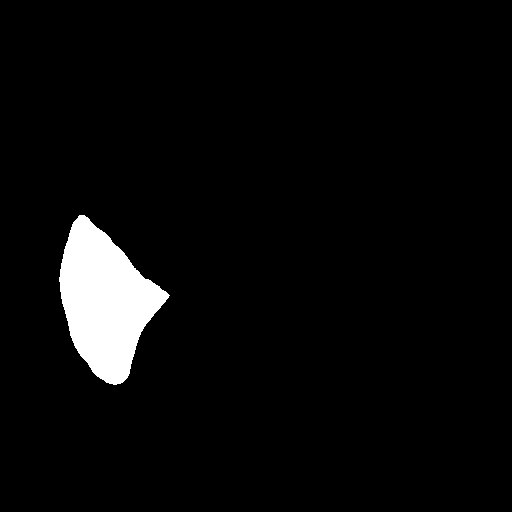} &
    \includegraphics[width =0.225\columnwidth]{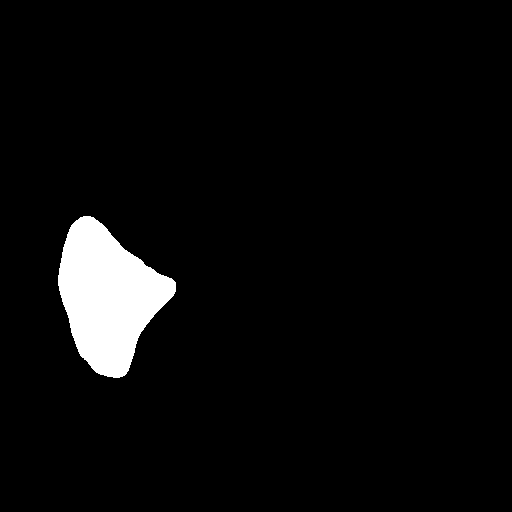} &
    \includegraphics[width =0.225\columnwidth]{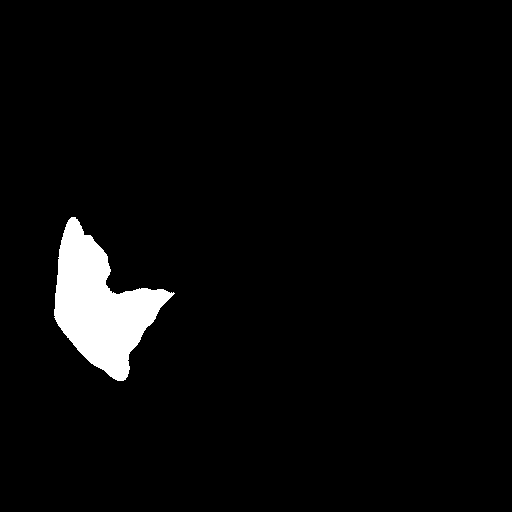} &
    \includegraphics[width =0.225\columnwidth]{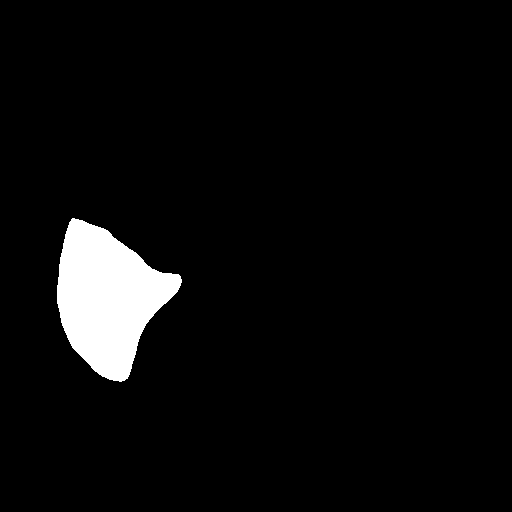} &
    \includegraphics[width =0.225\columnwidth]{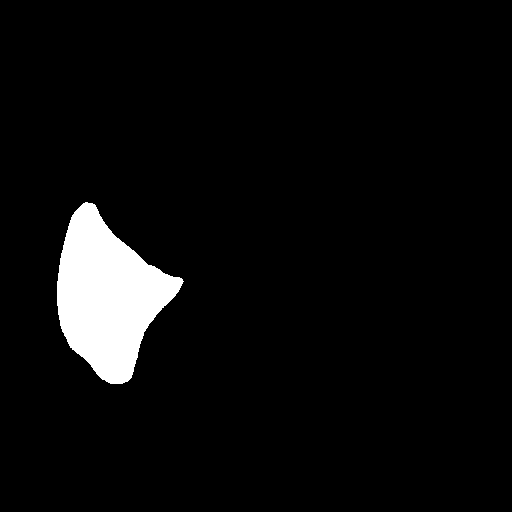} 
    \\
    \includegraphics[width =0.225\columnwidth]{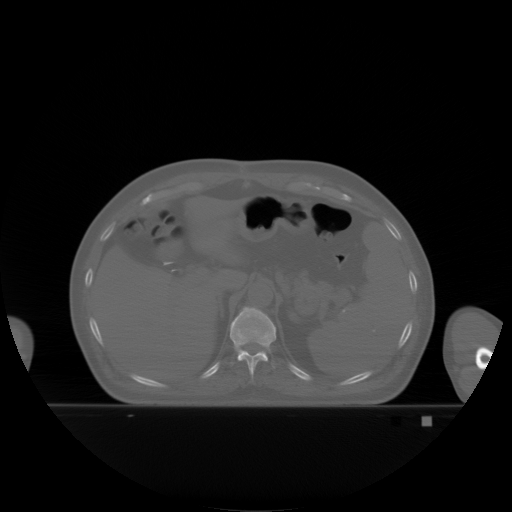} &
    \includegraphics[width =0.225\columnwidth]{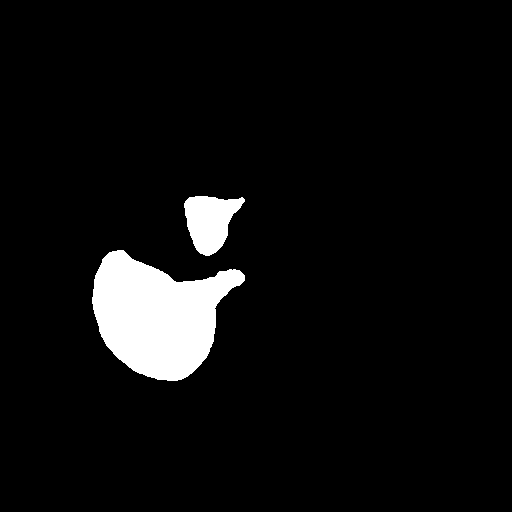} &
    \includegraphics[width =0.225\columnwidth]{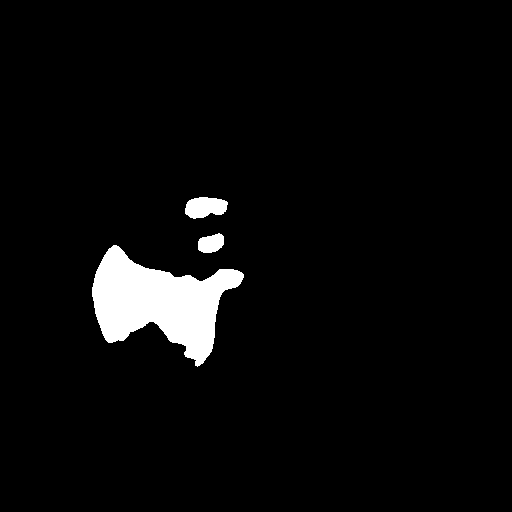} &
    \includegraphics[width =0.225\columnwidth]{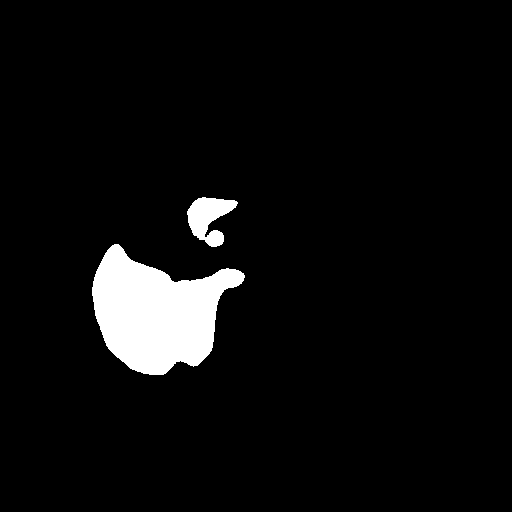} &
    \includegraphics[width =0.225\columnwidth]{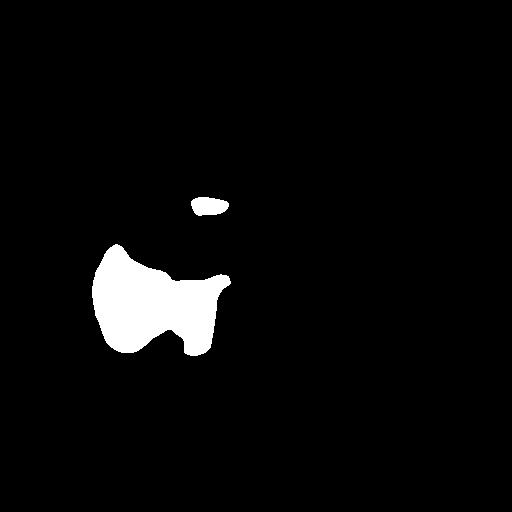}&
    \includegraphics[width =0.225\columnwidth]{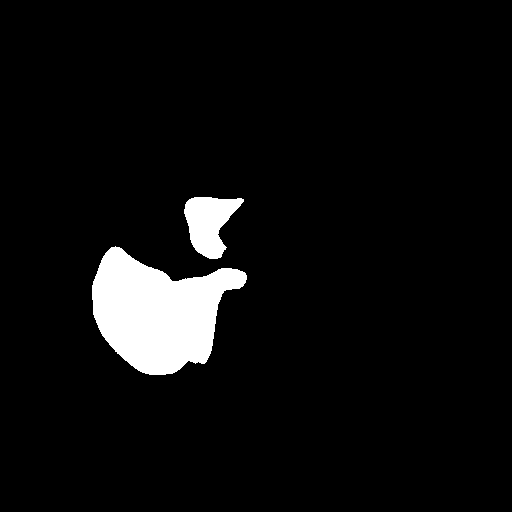}&
    \includegraphics[width =0.225\columnwidth]{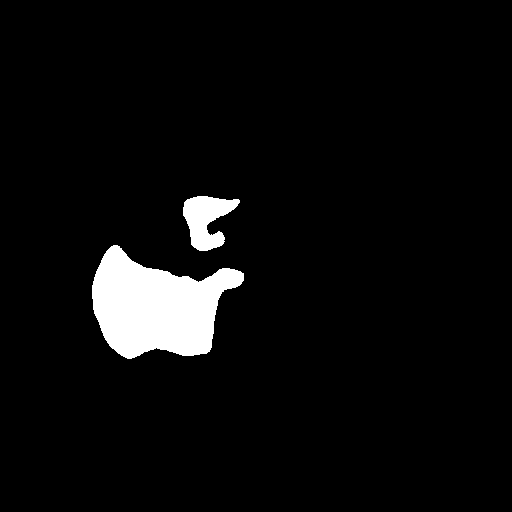} &
    \includegraphics[width =0.225\columnwidth]{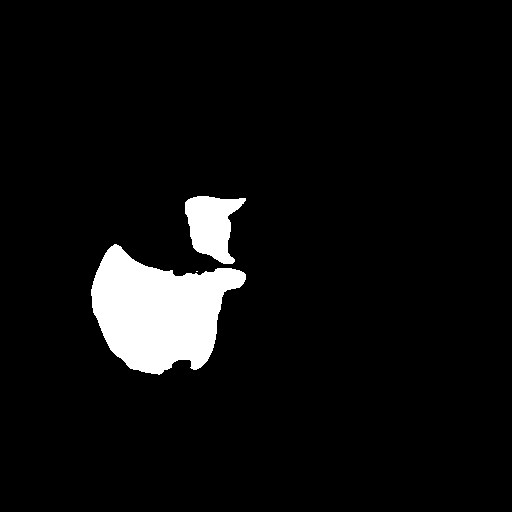} 
    \\
    \includegraphics[width =0.225\columnwidth]{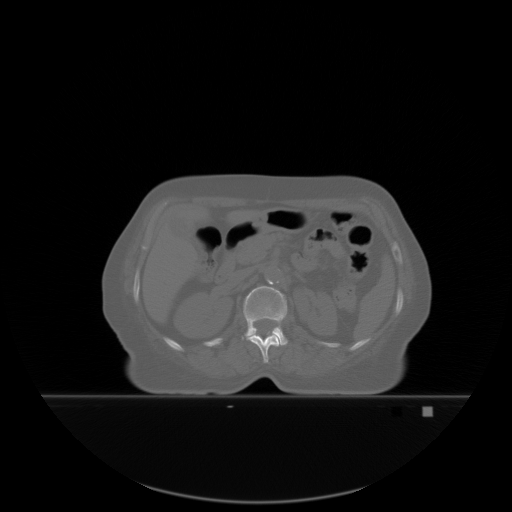} &
    \includegraphics[width =0.225\columnwidth]{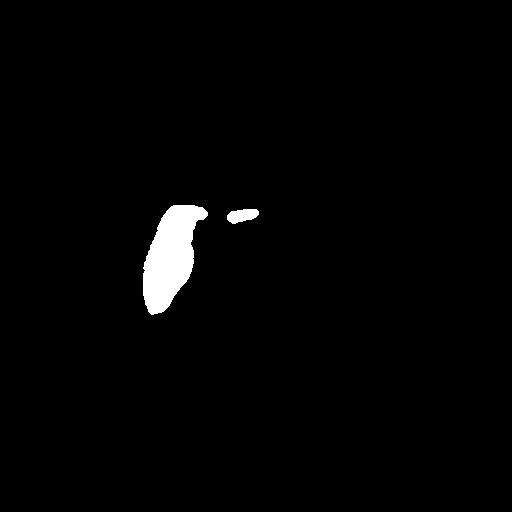} &
    \includegraphics[width =0.225\columnwidth]{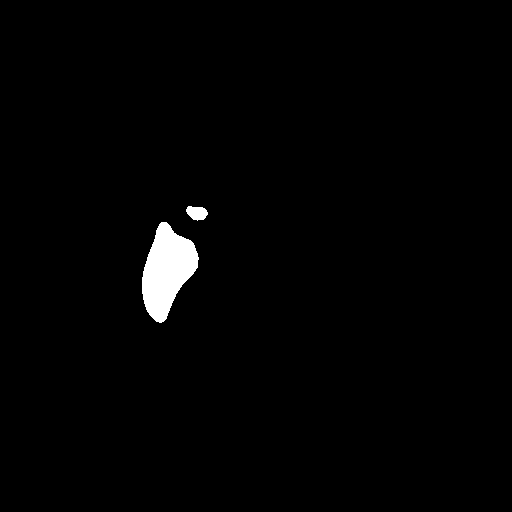} &
    \includegraphics[width =0.225\columnwidth]{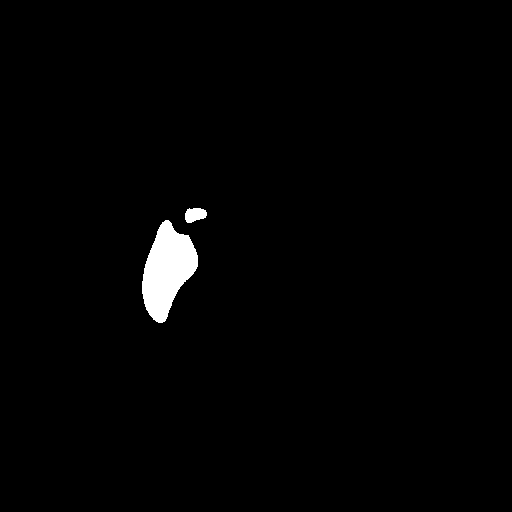} &
    \includegraphics[width =0.225\columnwidth]{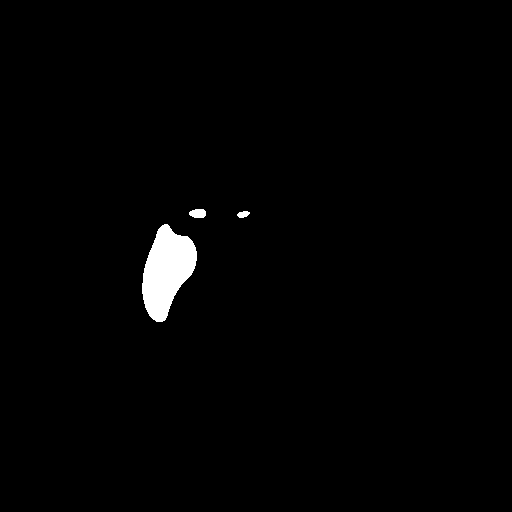} &
    \includegraphics[width =0.225\columnwidth]{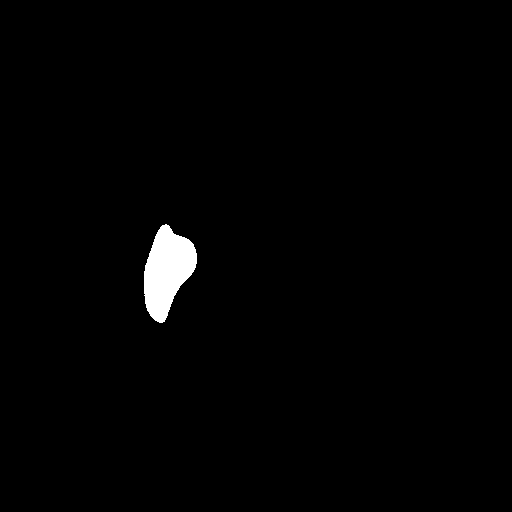} &
    \includegraphics[width =0.225\columnwidth]{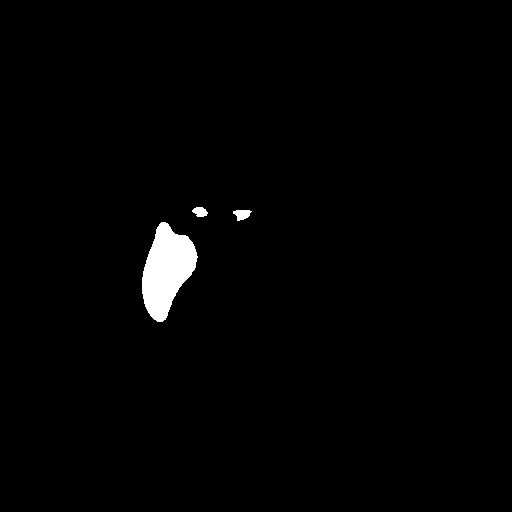} &
    \includegraphics[width =0.225\columnwidth]{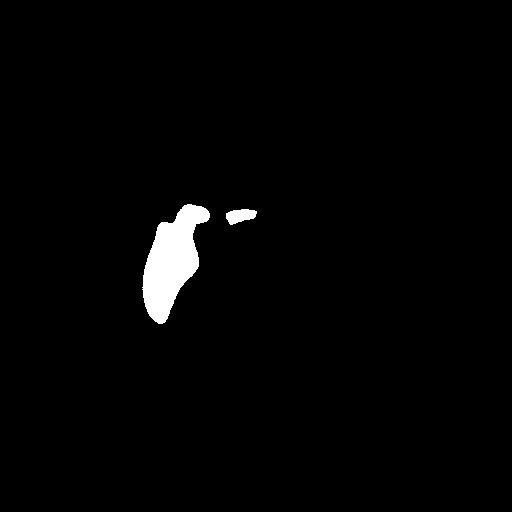} 
    \\
     (a) & (b) & (c) & (d) & (e) & (f) & (g) & (h) \vspace{-0.4cm}
    \end{tabular}
    \end{center}
    \caption{For the UNet architecture, visual results obtained on exemplary test set images. (a) CT images. (b) Ground truth maps. Results obtained by (c) BaselineMethod~\cite{ronneberger2015u}, (d) HausdorffDistanceMethod~\cite{karimi2019reducing}, (e) ActiveContourMethod~\cite{chen2019learning}, (f) CascadedFourierNet~\cite{cansizFourierNetShapePreservingNetwork2023}, (g) NonAdaptiveFourierLoss, which uses the proposed loss without adaptively updating the coefficients $\omega_n$, and (h) \textit{FourierLoss}, which uses the proposed loss with trainable coefficients $\omega_n$.}
    \label{fig:visual_results_unet}
\end{figure*}

\begin{figure*}
\begin{center}
\begin{tabular}{@{~}c@{~}c@{~}c@{~}c@{~}c@{~}c@{~}c@{~}}
\includegraphics[width =0.22\columnwidth]{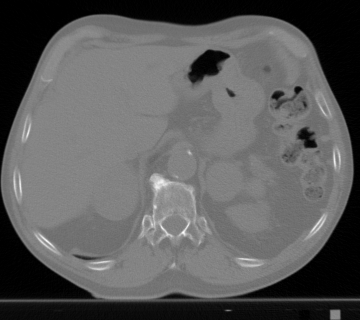} &
\includegraphics[width =0.22\columnwidth]{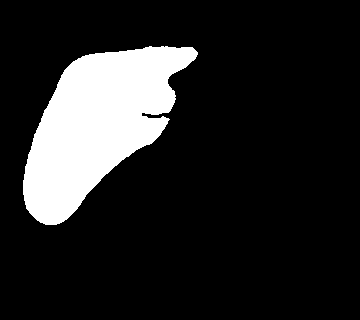} &
\includegraphics[width =0.22\columnwidth]{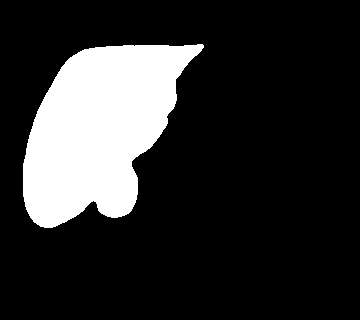} &
\includegraphics[width =0.22\columnwidth]{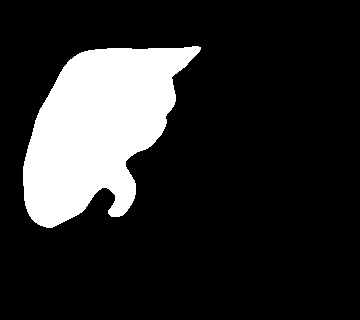} &
\includegraphics[width =0.22\columnwidth]{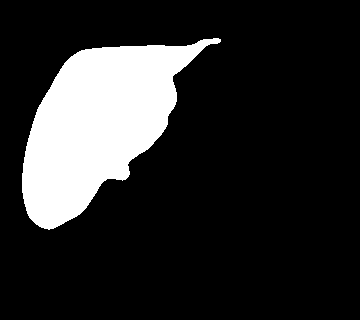} &
\includegraphics[width =0.22\columnwidth]{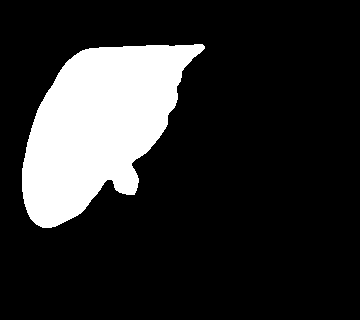} &
\includegraphics[width =0.22\columnwidth]{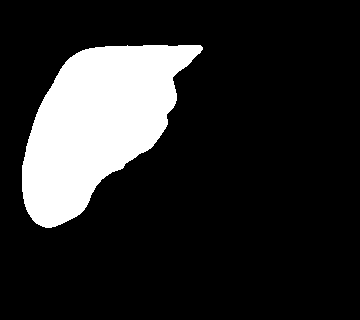} 
\\
\includegraphics[width =0.22\columnwidth]{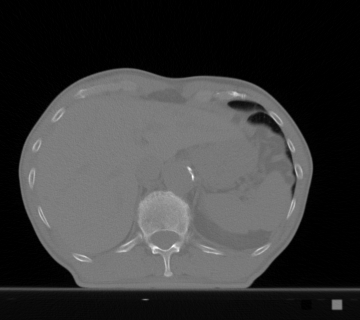} &
\includegraphics[width =0.22\columnwidth]{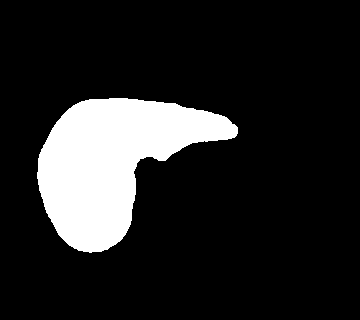} &
\includegraphics[width =0.22\columnwidth]{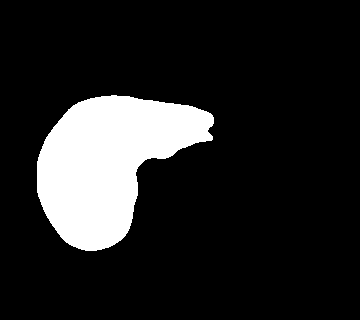} &
\includegraphics[width =0.22\columnwidth]{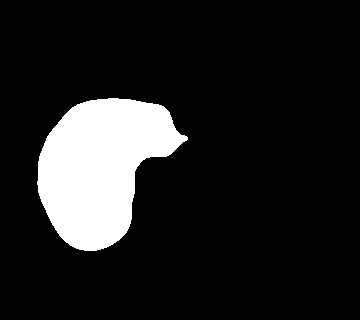} &
\includegraphics[width =0.22\columnwidth]{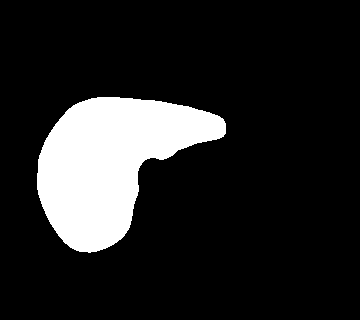} &
\includegraphics[width =0.22\columnwidth]{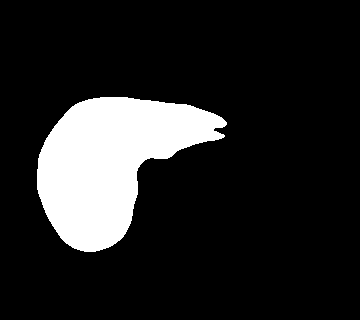} &
\includegraphics[width =0.22\columnwidth]{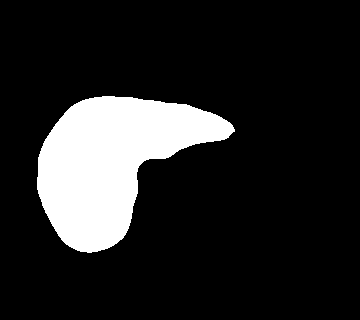} 
\\
\includegraphics[width =0.22\columnwidth]{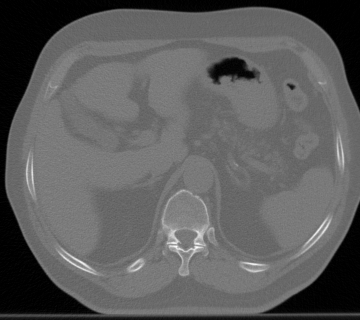} &
\includegraphics[width =0.22\columnwidth]{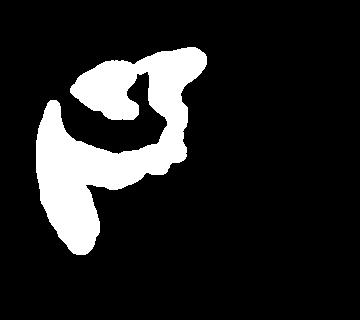} &
\includegraphics[width =0.22\columnwidth]{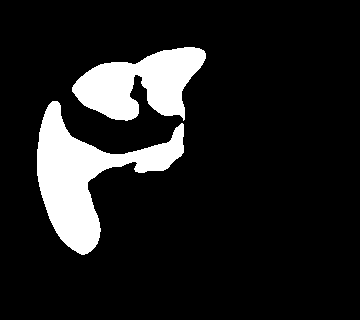} &
\includegraphics[width =0.22\columnwidth]{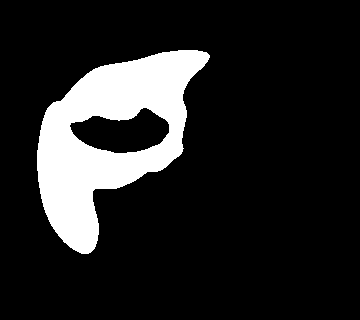} &
\includegraphics[width =0.22\columnwidth]{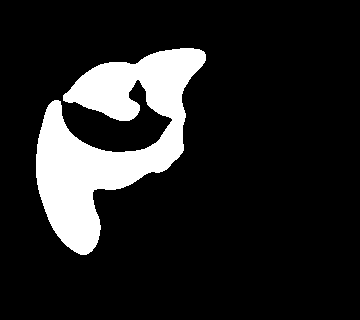}&
\includegraphics[width =0.22\columnwidth]{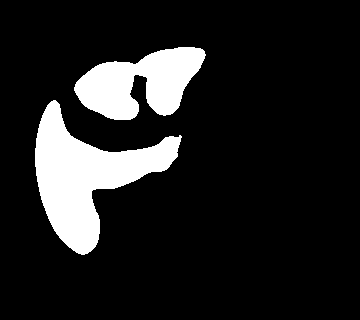} &
\includegraphics[width =0.22\columnwidth]{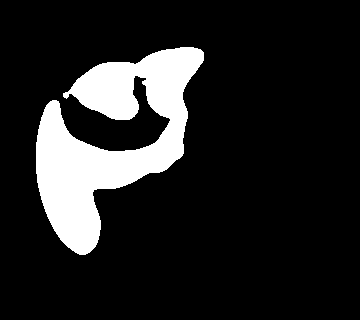} 
\\
\includegraphics[width =0.22\columnwidth]{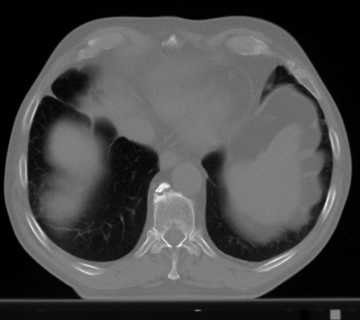} &
\includegraphics[width =0.22\columnwidth]{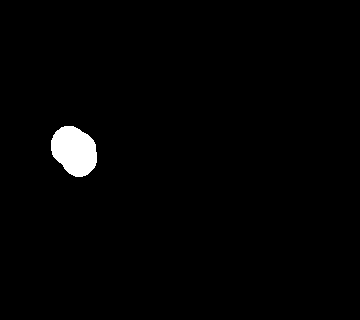} &
\includegraphics[width =0.22\columnwidth]{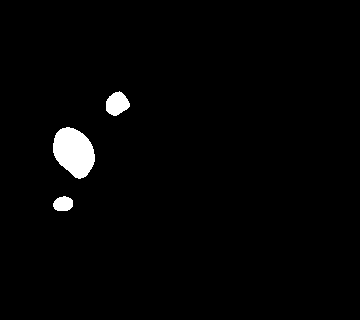} &
\includegraphics[width =0.22\columnwidth]{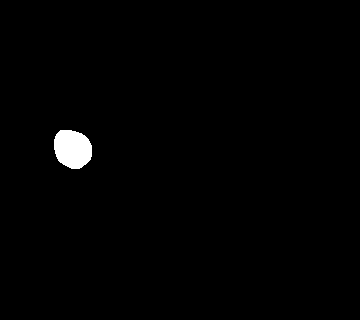} &
\includegraphics[width =0.22\columnwidth]{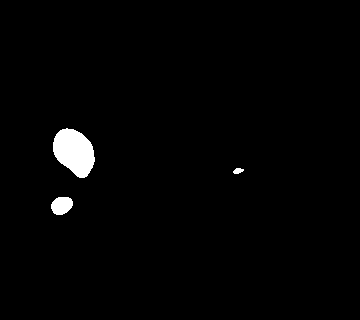}&
\includegraphics[width =0.22\columnwidth]{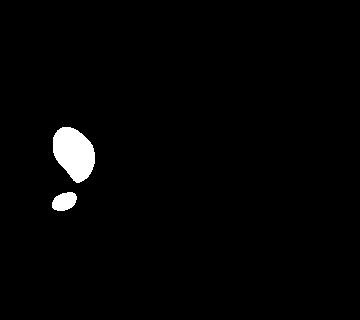} &
\includegraphics[width =0.22\columnwidth]{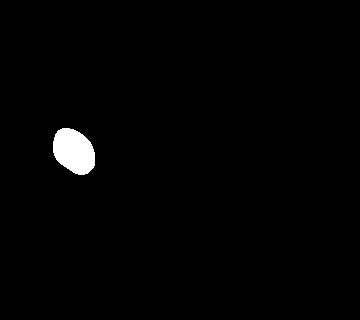}
\\
\includegraphics[width =0.22\columnwidth]{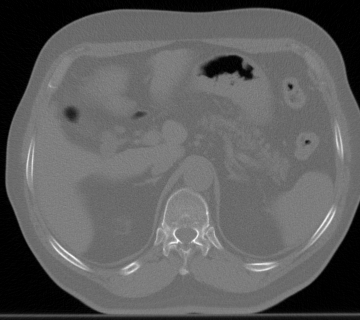} &
\includegraphics[width =0.22\columnwidth]{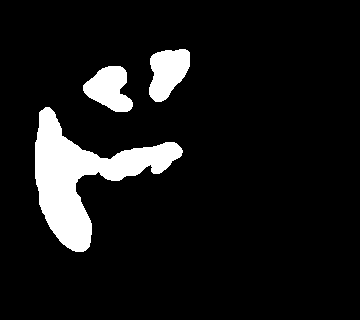} &
\includegraphics[width =0.22\columnwidth]{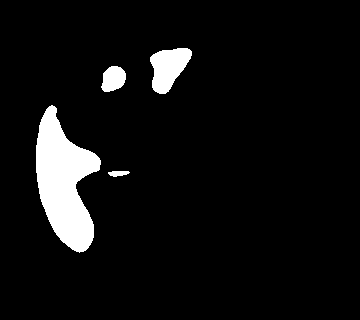} &
\includegraphics[width =0.22\columnwidth]{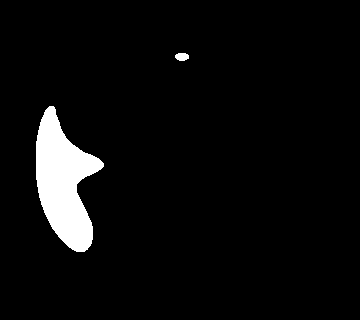} &
\includegraphics[width =0.22\columnwidth]{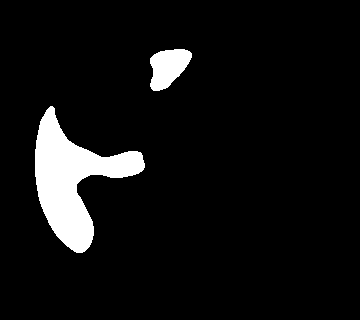}   &
\includegraphics[width =0.22\columnwidth]{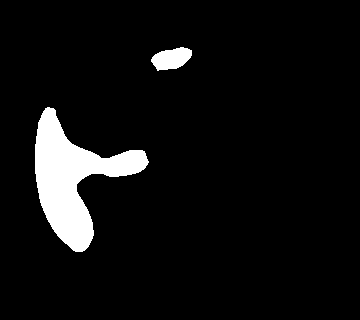} &
\includegraphics[width =0.22\columnwidth]{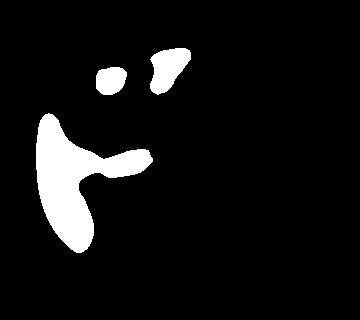} 
\\
 (a) & (b) & (c) & (d) & (e) & (f) & (g) \vspace{-0.4cm}
\end{tabular}
\end{center}
\caption{For the TransUnet architecture, visual results obtained on exemplary test set images. (a) CT images. (b) Ground truth maps. Results obtained by (c) BaselineMethod~\cite{chen2021transunet}, (d) HausdorffDistanceMethod~\cite{karimi2019reducing}, (e) ActiveContourMethod~\cite{chen2019learning}, (f) NonAdaptiveFourierLoss, which uses the proposed loss without adaptively updating the coefficients $\omega_n$, and (g) \textit{FourierLoss}, which uses the proposed loss with trainable coefficients $\omega_n$. Note that these visuals were on the cropped images; they were cropped just for better visualization whereas the original images were used in the experiments.}
\label{transunet_visual}
\end{figure*}

\section{Results and Discussion}

The quantitative test set results obtained by using the proposed \textit{FourierLoss} function with the UNet and TransUnet architectures are reported in Tables~\ref{table:Unet-results} and~\ref{table:Transunet-results}, respectively. These results were obtained by selecting the number $N$ of the Fourier descriptors as 2. The coefficients $\omega_n$ of these first two descriptors were selected as [3.0, 1.0] for the UNet architecture. We started with a larger coefficient for the first descriptor as UNet is a simpler network and we wanted to explicitly force this simple model to give more emphasis to learn the coarse outline especially in the first training epochs. On the other hand, they were selected as [1.0, 1.0] for the TransUnet architecture since this explicit enforcement is not necessary for this network due to the performance betterment achieved by its attention mechanism. The effects of selecting $N$ and the initial values of $\omega_n$ on the segmentation performance will be further investigated for the UNet architecture in Section~\ref{ablation_study}. Here it is worth noting that although we explicitly initialized the coefficients $\omega_n$ at the beginning, they were adaptively modified throughout network training to minimize the \textit{FourierLoss} function by backpropagation. Also note that different than the proposed model, the \textit{NonAdaptiveFourierLoss} method started with the same coefficients but did not modify them during training. 

Tables~\ref{table:Unet-results} and~\ref{table:Transunet-results} revealed that the use of the proposed \textit{FourierLoss} led to the best f-score, IoU, and Hausdorff distance both for the UNet and TransUnet architectures. Here it is also worth noting that the \textit{FourierLoss} did not give only the best average scores but also the lowest standard deviations, indicating the steadiness of the network convergence with the proposed loss function. We also applied the paired-sample t-test on the results. This statistical test showed that these improvements were statistically significant with $p < 0.05$.

For the UNet architecture, all methods improved the performance of the baseline, as shown in Table~\ref{table:Unet-results}. This is attributed to the fact that this architecture may require more refinement on its predictions as it is relatively simple, and the use of shape information in different ways resulted in performance improvement. The visual results given in Figure~\ref{fig:visual_results_unet} are also consistent with this fact. On the other hand, even though all methods improved the results, the improvement obtained by the \textit{HausdorffDistanceMethod} and \textit{ActiveContourMethod} was smaller compared to the three methods that use the Fourier descriptors in their design either by making architectural changes or defining a loss function. The reason may be as follows: when the predicted regions tend to be smaller in area than the ground truth regions (i.e., there exist false negative pixels), every method attempts to correct these false negatives thanks to their shape awareness. On the other hand, our experiments revealed that the loss functions of the \textit{HausdorffDistanceMethod} and \textit{ActiveContourMethod} might not be that effective for our problem, resulting in smaller liver regions at the end. This can also be observed in their precision and recall metrics. Since they tended to predict smaller regions, they yielded more false negatives but less false positives, which resulted in lower recall but higher precision metrics than the three methods using Fourier descriptors. Nevertheless, their f-scores, which is a trade-off between precision and recall, was lower than these three methods.

Moreover, for the UNet architecture, our model led to statistically significantly better ($p < 0.05$) quantitative results compared to the \textit{CascadedFourierNet} and \textit{NonAdaptiveFourierLoss} methods, which also used Fourier descriptors to quantify shape dissimilarity (Table~\ref{table:Unet-results}). The first comparison indicated that it was more effective to make use of the Fourier descriptors in defining a loss function rather than generating regression maps, learning of which were used as auxiliary intermediate tasks in a cascaded network. In other words, it indicated the effectiveness of forcing the network to learn the shape by using the proposed shape-aware loss function instead of employing auxiliary shape-related tasks. The comparison with the use of \textit{NonAdaptiveFourierLoss} revealed that the proposed loss update mechanism, which allowed to concurrently learn the coefficients $\omega_n$ of the Fourier descriptors in the loss function during network training, resulted in more accurate segmentations. The visual results were also consistent with this quantitative comparison. Figure~\ref{fig:visual_results_unet} shows that the proposed model estimated the shape outline as well as the details of the contour pixels better, thanks to its ability of modeling different levels of shape details and adapting itself to learn the importance of attending the correct detail level(s) during network training. This ability helped preserve the contour details, which in turn gave more accurate segmentations than the other methods especially towards the contour pixels. 

The TransUnet architecture includes transformer layers, which increase the complexity and learning capacity of the network. Thus, it has become one of the most successful segmentation models for various computer vision tasks. Nevertheless, we included this network architecture in our experiments to explore the effects of using the proposed \textit{FourierLoss} function to train a more powerful network. Both quantitative (Table~\ref{table:Transunet-results}) and visual (Figure~\ref{transunet_visual}) results showed that the other comparison methods could not outperform the baseline due to the increased learning power of the baseline. On the other hand, the proposed \textit{FourierLoss} function led to statistically significantly better results than the baseline ($p < 0.05$). Especially considering the results of the \textit{NonAdaptiveFourierLoss} method, this betterment could be attributed to the proposed adaptive loss update mechanism, which allows to simultaneously learn the Fourier loss coefficients $\omega_n$ by backpropagation. As seen in the first three rows of Figure~\ref{transunet_visual}, the proposed \textit{FourierLoss} function helped refine the shape outline by improving the results for the shape contour details. It was also effective to eliminate false positive pixels (the fourth row of Figure~\ref{transunet_visual}), increasing the precision. At the same time, it was successful to complete the missing parts of the shape (the last row of Figure~\ref{transunet_visual}), which resulted in obtaining less false negatives, and as a result, also increasing the recall. 

\begin{table*}[!ht]
\centering
\caption{For the UNet architecture, test set results obtained using the proposed \textit{FourierLoss} function with different number $N$ of the Fourier coefficients. Here the loss coefficient $\omega_1$ of the first descriptor was set to 3.0, and those of the later descriptors were set to 1.0.}
  \label{table:num_weight}
  \begin{tabular}{cccccc}
    \toprule
~~$N$~~ & Precision & Recall & F-score & IoU & Hausdorff d. \\ \hline  
1   & 88.85~$\pm$~1.42 & 93.14~$\pm$~0.76 & 90.12~$\pm$~0.61 & 84.36~$\pm$~1.10 & 4.71~$\pm$~0.18 \\
2   & 90.29~$\pm$~0.65 & 93.24~$\pm$~0.70 & 91.03~$\pm$~0.21 & 85.70~$\pm$~0.33 & 4.53~$\pm$~0.03 \\
3   & 89.56~$\pm$~1.60 & 92.84~$\pm$~1.50 & 90.43~$\pm$~0.54 & 84.96~$\pm$~0.82 & 4.57~$\pm$~0.10 \\
4   & 89.97~$\pm$~1.47 & 91.55~$\pm$~1.91 & 89.47~$\pm$~1.40 & 83.94~$\pm$~1.55 & 4.66~$\pm$~0.12 \\
5   & 88.59~$\pm$~2.29 & 93.44~$\pm$~1.69 & 90.11~$\pm$~1.03 & 84.32~$\pm$~1.47 & 4.71~$\pm$~0.20 \\ 
\bottomrule
\end{tabular}
\end{table*}

\begin{table*}[!ht]
\centering
\caption{For the UNet architecture, test set results obtained using the proposed \textit{FourierLoss} function with different initializations of the Fourier loss coefficients $\omega_n$. Here the number $N$ of the Fourier descriptors was set to 2.}
  \label{table:weight_init}
  \begin{tabular}{cccccc}
    \toprule
~~$[\omega_1,~\omega_2]$~~ & Precision & Recall & F-score & IoU & Hausdorff d. \\ \hline  
$[1.0,~1.0]$ & 89.18~$\pm$~2.05 & 91.84~$\pm$~2.53 & 89.53~$\pm$~0.41 & 83.76~$\pm$~0.73 & 4.70~$\pm$~0.12 \\
$[2.0,~1.0]$ & 88.25~$\pm$~1.36 & 94.20~$\pm$~1.36 & 90.35~$\pm$~0.31 & 84.67~$\pm$~0.51 & 4.71~$\pm$~0.11 \\
$[3.0,~1.0]$ & 90.29~$\pm$~0.65 & 93.24~$\pm$~0.70 & 91.03~$\pm$~0.21 & 85.70~$\pm$~0.33 & 4.53~$\pm$~0.03 \\
$[4.0,~1.0]$ & 88.46~$\pm$~0.86 & 93.71~$\pm$~0.72 & 90.20~$\pm$~0.37 & 84.59~$\pm$~0.48 & 4.64~$\pm$~0.07 \\
$[8.0,~1.0]$ & 88.41~$\pm$~1.88 & 92.46~$\pm$~2.86 & 89.27~$\pm$~1.51 & 83.51~$\pm$~1.61 & 4.77~$\pm$~0.13 \\
\bottomrule
\end{tabular}
\end{table*}

\subsection{Sensitivity Analysis}
\label{ablation_study}

The proposed \textit{FourierLoss} has two external parameters: the number $N$ of the Fourier descriptors used to quantify the shape and the initial values of the Fourier loss coefficients $\omega_n$. Next, we investigated the effects of their selections to the segmentation performance for the case where we used the UNet architecture. For this case, we selected $N = 2$ and  $[\omega_1,~\omega_2] = [3.0,~1.0]$. To this end, for each parameter, we fixed the selected value of the other and measured the test set performance metrics as a function of the parameter of interest. Note that to make fair comparisons, for the corresponding runs, we initialized the network weights using the same seed in the random number generation and processes of Pytorch. 

In general, a function can be reconstructed from the coefficients in its Fourier series when the number $N$ of the coefficients goes to infinity. The first Fourier coefficient gives the general outline of this function whereas the latter ones add more and finer details. In our case, this corresponds to modeling the general shape outline by the first Fourier coefficient and the finer details of the shape contours by the latter ones (see Figure\ref{fig:fdmap}). Selecting $N$ too small may cause not to model the details, and thus, an inadequate representation of the shape. On the other hand, selecting $N$ too big may result in representing noise in the shape contours, causing to overfit the contours. In our experiments, using the first two Fourier descriptors, one representing the outline and the other the details, was sufficient to obtain accurate results. As shown in Table~\ref{table:num_weight}, the network could not learn the shape details when $N=1$, and most probably started learning redundant details when $N > 3$. Note that in this analysis, as consistent with our previous selection, we set the loss coefficient $\omega_1$ of the first Fourier descriptor to 3.0 and those of the latter descriptors to 1.0, giving the same initial importance to learning different levels of the shape detail except the general outline. 

In the \textit{FourierLoss} definition, the $\beta_I$ penalty term is defined as a linear combination of the differences between the Fourier descriptors of the ground truth objects and their corresponding objects in the prediction maps. The coefficient $\omega_n$ determines the importance of the $n$-th descriptor in this linear combination. Since $N = 2$, $\omega_1$ defined for the first Fourier descriptor gives information about the coarse outline, and $\omega_2$ defined for the second descriptor emphasizes the contour details. For the UNet architecture, we initialized these coefficients such that $\omega_1 \ge \omega_2$ since we wanted to explicitly enforce the network to learn the coarse outlines, especially in the first epochs; the network may dynamically change its attention to learn the contour details in the later epochs using the proposed adaptive loss update mechanism. Since these coefficients are used in a linear combination, the ratios between them, rather than their absolute values, determines the relative impact of each Fourier descriptor on the loss function. Thus, in this sensitivity analysis, we fixed $\omega_2 = 1.0$ and changed the value of $\omega_1$ from 1.0 to 8.0, as reported in Table~\ref{table:weight_init}. This table shows that relatively small or big values of $\omega_1$ decreased the segmentation performance. This may be because of the following. For the former case, the network may not learn the general outline adequately, and thus, learning the shape details without learning the general outline could not help. For the latter case, the network may lose the shape details by merely focusing on the first Fourier descriptor as a result of using too big initial values for the coefficient $\omega_1$. In our application, we set the balance between these two factors by selecting $\omega_1 = 3.0$. Of course, this balance depends on the network and dataset characteristics. It is worth noting that all results obtained in this sensitivity analysis, by using different parameter selections, outperformed the performance of the \textit{BaselineMethod} in the f-score and IoU metrics, indicating the effectiveness and contribution of the proposed shape-aware loss function. The thorough analysis of this parameter for the other types of networks and datasets and using alternative ways of combining differences between different Fourier coefficients in the $\beta_I$ definition are considered as future research work. 

\section{Conclusion}
In this work, we proposed a novel shape-aware loss function, \textit{FourierLoss}, that can be integrated into the training of any encoder-decoder network architecture. This loss function relied on using the Fourier descriptors to quantify the shape dissimilarity of the objects in the ground truth and predicted segmentation maps, and penalizing this shape dissimilarity in network training. Unlike the previous methods, the proposed \textit{FourierLoss} function enabled to dynamically determine how much shape details were utilized in each epoch of the training process. This was achieved with the adaptive loss update mechanism that adjusted the coefficients of the Fourier descriptors, each of which corresponded to a different level of the shape detail, in the loss function by backpropagation. Our experiments on 2879 CT images of 93 subjects demonstrated that this proposed shape-aware loss function improved the segmentation performance of the baseline method, for two different network architectures, as well as those that incorporated alternative shape preserving loss functions into the baseline. In this work, we used the shape dissimilarity between the ground truth and the predicted segmentation maps to define a penalty term for the cross-entropy loss. This penalty term was defined for each image separately, and used for every pixel in the same image. As future work, one may use this penalty term for false negative and false positive pixels only, forcing the network to more attend to correcting its incorrectly predicted pixels.

\section*{Acknowledgments}
This work was supported by the Scientific and Technological Research Council of Turkey, grant no: TUBITAK 120E497.



 \bibliographystyle{elsarticle-harv} 
 \bibliography{whole}


%
\end{document}